\UseRawInputEncoding
\documentclass[conference]{IEEEtran}
\IEEEoverridecommandlockouts
\usepackage{cite}
\usepackage{amsmath,amssymb,amsfonts}
\usepackage{algorithmic}
\usepackage{graphicx}
\usepackage{textcomp}
\usepackage{xcolor}
\def\BibTeX{{\rm B\kern-.05em{\sc i\kern-.025em b}\kern-.08em
    T\kern-.1667em\lower.7ex\hbox{E}\kern-.125emX}}

\usepackage{algorithmic}
\usepackage{algorithm}

\usepackage{booktabs}
\usepackage{multirow}

\begin{document}

\title{An Emergency Disposal Decision-making Method with Human--Machine Collaboration*\\
{\footnotesize \textsuperscript{*}Note: Sub-titles are not captured in Xplore and
should not be used}
\thanks{Identify applicable funding agency here. If none, delete this.}
}
\author{\IEEEauthorblockN{1\textsuperscript{st} Yibo Guo}
	\IEEEauthorblockA{\textit{School of Computer and Artificial Intelligence} \\
		\textit{Zhengzhou University}\\
		ZhengZhou, China \\
		ieybguo@zzu.edu.cn}
	\\
	\IEEEauthorblockN{3\textsuperscript{rd} Yingkang Zhang}
	\IEEEauthorblockA{\textit{Shool of Computer and Artificial Intelligence} \\
		\textit{Zhengzhou University}\\
		ZhengZhou, China \\
		zhangyingkang@gs.zzu.edu.cn}

	\and
	
		\IEEEauthorblockN{2\textsuperscript{nd} Jingyi Xue}
	\IEEEauthorblockA{\textit{School of Computer and Artificial Intelligence} \\
		\textit{Zhengzhou University}\\
		ZhengZhou, China \\
		202022172013271@gs.zzu.edu.cn}
	
	\\
	\IEEEauthorblockN{4\textsuperscript{th} Mingliang Xu$^{\ast}$}
	\IEEEauthorblockA{\textit{School of Computer and Artificial Intelligence} \\
		\textit{Zhengzhou University}\\
		ZhengZhou, China \\
				}
}

\maketitle

\begin{abstract}
Rapid developments in artificial intelligence technology have led to unmanned systems replacing human beings in many fields requiring high-precision predictions and decisions. In modern operational environments, all job plans are affected by emergency events such as equipment failures and resource shortages, making a quick resolution critical. The use of unmanned systems to assist decision-making can improve resolution efficiency, but their decision-making is not interpretable and may make the wrong decisions. Current unmanned systems require human supervision and control. Based on this, we propose a collaborative human--machine method for resolving unplanned events using two phases: task filtering and task scheduling.

In the task filtering phase, we propose a human--machine collaborative decision-making algorithm for dynamic tasks. The GACRNN model is used to predict the state of the job nodes, locate the key nodes, and generate a machine-predicted resolution task list. A human decision-maker supervises the list in real time and modifies and confirms the machine-predicted list through the human--machine interface. In the task scheduling phase, we propose a scheduling algorithm that integrates human experience constraints. The steps to resolve an event are inserted into the normal job sequence to schedule the resolution. We propose several human--machine collaboration methods in each phase to generate steps to resolve an unplanned event while minimizing the impact on the original job plan. 
\end{abstract}


\section{Introduction}
Expanding use of human--computer interaction technology means that artificial intelligence systems are increasingly being used in unmanned warehouses, smart factories, and other work environments to solve complex tasks. Complex tasks have potential risks \cite{b1} and are prone to unplanned interruptions, as with machine failures in unmanned warehouses or material fires in factories. Failure to resolve and control such events in a timely manner often leads to delays or even serious accidents. The key to preventing such dangers is the formulation of event resolution decisions. Prompt decisions can minimize the losses caused by these events. Researchers have been studying emergency disposal and response for several decades. Emergency problems are considered to be temporal and spatial decisions with uncertainty \cite{b2}. Exploring these decisions has garnered significant research interest.

In recent years, many scholars have proposed the importance of critical nodes in emergency disposal and response \cite{b3, b4, b5, b6}. Experimental evidence has shown that specific emergency disposal measures for these critical nodes can reduce the negative impact of emergencies. However, most emergency scheduling methods focus on emergency resource allocation \cite{b7, b8} and material path optimization \cite{b9, b10} for large-scale disasters. We focus on small-scale emergency events in environments with strict operational procedures. In particular, we emphasize the coordination and arrangement of emergency disposal tasks within normal functions, which can be abstracted as an insertion problem of new tasks. Therefore, we divide emergency disposal into two stages: task filtering and task scheduling. In the first stage, we filter out the key nodes. In the second stage, we address the insertion of emergency disposal tasks into the normal sequence of work.

Unmanned systems excel at logical reasoning, large-scale data computation, and storage, which can improve computational efficiency and provide quick suggestions. However, machines are prone to decision-making risks and system loss of control when facing complex conditions. When the suggestions made by the system contradict the decisions made by human decision-makers, it can further reduce trust in the system. Emergency disposal requires unmanned systems to improve decision-making efficiency, but, at present, unmanned systems cannot operate without human supervision and control.

In summary, we propose a method of human--machine collaboration to resolve unplanned events. Through the study of normal operational flow and  emergency disposal scenarios, we divide the process into two phases: task filtering and task scheduling. We model the scenarios using graph computations, and we abstract the emergency disposal problem as a problem of key node prediction and graph node insertion. We summarize the specific research contributions of this article as follows.

\begin{itemize}
	\item We propose a human--machine collaborative decision-making algorithm for dynamic tasks integrating multiple features to predict the status of station nodes and listing nodes with likely events as key nodes. To address potential discrepancies between the algorithm and actual situations, we introduce human--machine collaboration to reduce the workload of human decision-makers while improving human trust in the system decisions.
	
	\item We propose an emergency disposal task scheduling algorithm using  human experience and safe behavior constraints. The emergency disposal task list is obtained using a human--machine collaborative decision-making algorithm, with Double DQN used to solve the task scheduling problem. To ensure that the response actions are within safe constraints, an Action-Mask mechanism is used to filter the reinforcement learning action set based on human experience.
\end{itemize} 

\section{Related Work}
\subsection{Traditional Emergency Disposal and Response Research} Emergency disposal is a multi-stage process, with identifying key nodes for formulating the emergency disposal task list as the first step. N.~Khakzad et al. \cite{b11} introduced graph theory analysis into the emergency disposal problem in factories. By identifying the devices that play a key role in the accident initiation and propagation through graph centrality measurement, the corresponding resolution can be carried out, effectively reducing the impact of accidents. Ding et al.\cite{b4} proposed a spatio-temporal evolution simulation method using collaborative effects and accident evidence, visualizing the accident evolution and related risks, to identify the most vulnerable and dangerous units for guiding emergency disposal. Qi et al. \cite{b6} argued that emergency events are a two-way process of propagation. Using bidirectional connection graphs to visualize the interaction of hazards between multiple nodes and using the safety factor calculated on the graph as the basis for deploying safety barriers, the spread of emergency events can be reduced. They have all emphasized the importance of key positions in emergency disposal, mainly through methods such as event diffusion prediction and vulnerability calculation to predict and filter key nodes. Predicting key nodes essentially means predicting the state of those nodes, that is, predicting the spread of the emergency. The spread of events is a complex spatio-temporal problem with multiple features, and single-feature calculation methods may overlook important nodes.

After formulating the emergency disposal task list, it is necessary to schedule the tasks appropriately. We abstract this problem as one of inserting emergency disposal tasks into the original task order, making use of similar studies on the adaptive job-shop scheduling problem (JSSP) \cite{b12}. One solution \cite{b13, b14} used rescheduling strategies to solve the problem of plan failure caused by changes in arrival time, processing time, and machine failures. Similar online scheduling algorithms also exist \cite{b15}. Reinforcement learning fully reflects the characteristics of decision-making problems through state or action rewards, making it suitable for solving such scheduling problems. Wang et al. \cite{b16} designed a knowledge-based dynamic scheduling system based on multi-agent systems and proposed an adaptive scheduling strategy using weighted Q-learning that guides agents in dynamic environments. Shahrabi et al. \cite{b17} used reinforcement learning with a Q-factor algorithm to enhance the performance of dynamic job shop scheduling methods that considers the arrival of random tasks and machine failures. However, these only consider the impact of failures on normal operations and do not consider the insertion of emergency disposal tasks.

\subsection{Human--Machine Collaborative Methods} 
In recent years, human--machine collaborative decision making has gained significant interest. Research has proceeded in roughly four directions: human--machine energetic assignment, intelligent learning and modification, contextual adaption, and interaction patterns. Intelligent learning and correction methods can be further divided into learning from human demonstration (LFD), learning from human intervention (LFI), and learning from human evaluation (LFE) according to the human--robot interaction. LFD is also known as imitation learning (IL), where autonomous systems learn to perform a task by observing human demonstrations \cite{b18}. However, LFD relies on a large dataset of expert demonstrations, which is difficult to generalize for new scenarios and situations not available in the training set. LFI improves the accuracy or efficiency of machine learning models through human intervention, error correction, and guidance during the learning process of the system. Saunders et al. \cite{b1} proposed a framework for safe reinforcement learning that introduces human intervention, with humans stopping dangerous and undesirable behaviors by providing corrective feedback or controlling the machine. However, such an approach is not suitable for solving complex problems with high-dimensional state spaces with insufficient samples. J.~Spencer et al. \cite{b19} argued that LFI can compensate for IL's complete reliance on non-strategic human presentation and the drawbacks of time-consuming and burdensome human involvement. There are already the well-researched LFI frameworks TAMER \cite{b20} and COACH \cite{b21}. The main difference between LFE and LFI is that LFI receives feedback during the learning process, whereas LFE receives feedback after the learning process has occurred. To address the problems in these single-interaction modality frameworks, research in recent years has tended to combine single human--machine collaborative modalities. N.~R.~Waytowich et al. \cite{b22} proposed a Cycle-of-Learning for Autonomous Systems (CoL) that combines LFD, LFI, and LFE. The CoL conceptual framework integrates three modes of interaction and defines criteria for switching between different interaction modes. V.~G.~Goecks et al. \cite{b23} implemented a combined LFD and LDI for real-time training of autonomous systems and demonstrated in experiments that the combined framework significantly improved the efficiency and safety of system learning over a single interaction approach. In this paper, we use different human--computer collaboration methods  in the two phases of task filtering and scheduling according to different requirements.

\section{Problem Descriptions}
In this section, we define the emergency events that our paper addresses. Based on a comprehensive analysis of work processes, work environments, emergency event characteristics, and emergency disposal, we use graph theory to model this information. We also introduce key concepts and formal definitions of emergency disposal. Finally, we define the general problem of emergency disposal.

\subsection{Concept Definition}
\textbf{Definition 1: Emergency Events} The emergency events addressed in this paper refer to small-scale unexpected events that occur outside the original work plan and affect the workflow in situations with strict work processes. Emergency events occur suddenly and are difficult to predict in advance. They develop over time and, when they reach a certain level, spread and have negative consequences on other processes. Therefore, emergency events are characterized by their suddenness, uncertainty, destructiveness, and diffusion.

\textbf{Definition 2: Work Station Graph} The work station graph in this article is an abstract weighted graph $G=\left(V_{P}, E_{P}, D_{E}\right)$ that represents and preserves important details of the physical space in the work environment. $V_P=\left\{v_1,v_2,\cdots,v_m\right\}$  represents the set of station nodes in the overall environment, abstracting each work area as a node. $E_{P}$ represents the road connectivity between different work areas. $d_{ij}\in{D_E}$ denotes the distance between nodes $v_i$ and $v_j$.

\textbf{Definition 3: Emergency Event Diffusion Graph} The emergency event diffusion graph is an undirected weighted graph $G^\prime=(V_P,E_S,D)$ generated from the work station graph. $V_P$ denotes the set of nodes in the operational area, which is the same as the set of station nodes in the work station graph. The set of edges $E_S$ represents the road connectivity relationship and the implied pipeline and circuit connectivity relationship between each operation region. The weights of station node $v_i$ are attributes represented by a tuple $T_i^P=(k_i,F_i^a,w_i)$, where $k_i$ and $F_i^a$ denote the state and the emergency event type of station node ${v}_{i}$, respectively. If the node is normal, $k_i=F_i^a=0$, with $k_i=1$ indicating the node is in the emergency event state and $k_i=2$ indicating the node has been damaged by the emergency event. $w_i$ denotes the resource value of the node. The value is cleared when $k_i=2$. $D\in{R}^{N\times N}$ is the adjacency matrix, $N=\left|V_P\right|$. The set of emergency event nodes $V^a=\left\{v_h,\cdots,v_j\right\}$ is a subset of $V_P$, with state $k=1$.

\textbf{Definition 4: Normal Work} Normal work refers to the originally planned arrangement of operations. The work plan at the station node is split into multiple processes. One
process is a normal work node. A work node is the smallest unit and cannot be split again. Taking unmanned storage as an example, the storage task after the AGV arrives at the designated storage location can be subdivided into the following work nodes: information confirmation, location verification, scanning and registration, positioning, and storage. We use $nw_i\in W_t^{\mathrm{work}}$ to denote the set of normal work nodes corresponding to station node $v_i$ at moment $t$, where $nw_i=\left\{v_i^1,v_i^2,\cdots,v_i^n, v_i^n\right\}$ denotes the $n$th process of station node $v_i$. $W_t^{\mathrm{work}}=\left\{nw_i,\cdots,nw_j\right\}$ is the list of normal work nodes of all station nodes. The normal work node $v_i^n$ corresponds to the attribute $T_{\mathrm{in\ }}^{\mathrm{normal\ }}=\left(et_{in},w_{\mathrm{in\ }}\right)$, which indicates the work deadline and work value in turn. The work needs to be completed before the deadline, or the work is delayed and a negative value will be assigned to the work value. The work nodes are added to the set of delayed work nodes $U$.

\textbf{Definition 5: Process Subgraph} We add the corresponding normal work nodes to the station nodes according to the actual operation requirements. All normal work nodes of a station node form a process subgraph, which is represented by $mg_i=\left(nw_i,e_i^{mg}\right)\in mG$, with $nw_i$ denoting all the work nodes corresponding to the station node and $e_i^{mg}$ denoting the order between them. Each subgraph possesses two virtual job vertices: a start node and an end node, indicating the start and end of that work process. The normal work process is abstracted as the addition of process subgraphs for station nodes according to the schedule, and when a work item is completed, the corresponding normal work nodes in the process subgraph are deleted and added to the set of completed work nodes $S^t$.

\textbf{Definition 6: Emergency Disposal Task} An emergency disposal task needs to be completed outside of the normal operational schedule to resolve emergency events that occur at station nodes and to protect nodes where emergency events have not yet occurred. The set of emergency disposal task nodes is $V^E=\left\{v_s^\ast,\cdots,v_t^\ast\right\}$, where $v_s^\ast$ denotes the emergency disposal task node of station node $v_s$. If $k_s=0$, the system performs a preventative-type task, such as setting firewalls in the connection channel to avoid node resources and operations from being damaged. If $k_s=1$, the system performs a rescue-type task. It should be noted that the set of emergency event nodes and the set of emergency disposal task nodes are not in a one-to-one correspondence, and the station nodes corresponding to emergency disposal task notes are both abnormal nodes that are experiencing emergency events and normal nodes that have not yet experienced emergency events. The weight of the emergency disposal task node $v_i^\ast$ is $T_i^e=\left(et_i,w_i\right)$, which is similar to the representation of normal work nodes and is represented as the task deadline and task value in order, where the task deadline is related to the node state, and the emergency disposal task deadline $et_i=0$ when $k_i=2$

\textbf{Definition 7: Emergency Disposal Situation Graph} An emergency disposal situation graph is a weighted graph $G^{\prime\prime}=\left(V,E\right)$ generated by adding process subgraphs on the basis of the work station graph. By adding and deleting work nodes or task nodes from the graph, we  directly reflect the normal operation and emergency disposal. When $V=V_P\cup V^n\cup V^e$, each node is distinguished by the identifier $\mathrm{type}$. $V_P$ denotes the station nodes, with node type identifier $\mathrm{type}=0$.   The normal work nodes $V^n$ have $\mathrm{type}=1$. The emergency disposal task nodes $V^e$ have $\mathrm{type}=2$. $E$ consists of the linkage relationships between station nodes, the subordination relationship, and the order relationship of the nodes.

\textbf{Definition 8: Machine Decision Set} The set ${Dec\_machine}=\left\{V^{m E}, O\right\}$ denotes the decisions made by the machine in the human--machine collaborative resolution, where $V^{mE}$ denotes the list of machine-generated emergency disposal tasks, and $O$ denotes the sequence of works and tasks for each station node given by the machine through the emergency disposal task scheduling algorithm.

\textbf{Definition 9 Human Decision Set} The set ${Dec\_human}=\left\{task\_size,V^{hE},check,constraint\_list\right\}$ represents the decisions made by humans, where $task\_size$ represents the task list window size, $V^{hE}$ represents the result of human corrections to the machine-generated emergency disposal task list; $check$ is a binary variable indicating whether the human approves the emergency disposal task list given by the machine; and $constraint\_list$ indicates the human-provided constraint list used to formulate the corresponding safety constraints.

\subsection{Emergency Disposal}

Emergency disposal refers to the determination and scheduling of the emergency disposal task list in the task filtering and scheduling phases, respectively. Emergency disposal limits the spread of the negative results of the emergency event to ensure normal operation.

We abstract the emergency disposal problem as follows. In the task filtering phase, given a graph $G^\prime=(V_P,E_S,D)$ that represents the spread of the emergency events, the emergency disposal task list $V^E=\left\{v_s^\ast,\cdots,v_t^\ast\right\}$ is determined by predicting the spread of the emergency and locating the key station nodes. In the task scheduling stage, the emergency disposal situation graph $G^{\prime\prime}$ adds the emergency disposal task node corresponding to the station node to the process subgraph, generating the work and task sequence $o_i\in O\left(t\right)$ of the node. Execute the emergency disposal task before the node's state develops to $k=2$, which makes the node restore k=0 state. Otherwise, the task fails, and the three graphs are updated accordingly: the resource value of the station node is cleared to zero ($w\gets0$); the task value in the process subgraph of the station node is assigned a negative value and added to the set of failed task nodes $U$; and the emergency event diffusion and disposal situation graphs with all connected edges of this station node are disconnected.

The goal of the emergency disposal problem is to maximize the maintenance of normal work steps and minimize the effect of emergencies:
\begin{equation}\label{eq1}
	T=\max \sum {Value}(O)+\min \sum {Value}\left(V^{a}\right),
\end{equation}
where ${Value}(O)$ indicates the completion of normal steps; and ${Value}\left(V^{a}\right)$  indicates the total number of nodes affected by emergency events.

\subsection{Diffusion of Emergency Events}
We also study the spread of emergency events that affect other station nodes. Taking fire as an example, a fire at one station node may spread from a road to an adjacent node or through a pipeline to a node that is not directly connected on the surface. The emergency diffusion graph can respond to the emergency before it spreads further, helping to eliminate the effects of the emergency quickly. Using this description of the diffusion of emergencies, we divide them into categories of space-dependent and time-dependent.

a) Spatial dependence

\begin{figure}[ht]
	\centering
	\includegraphics[width=9.0cm,height=3.3cm]{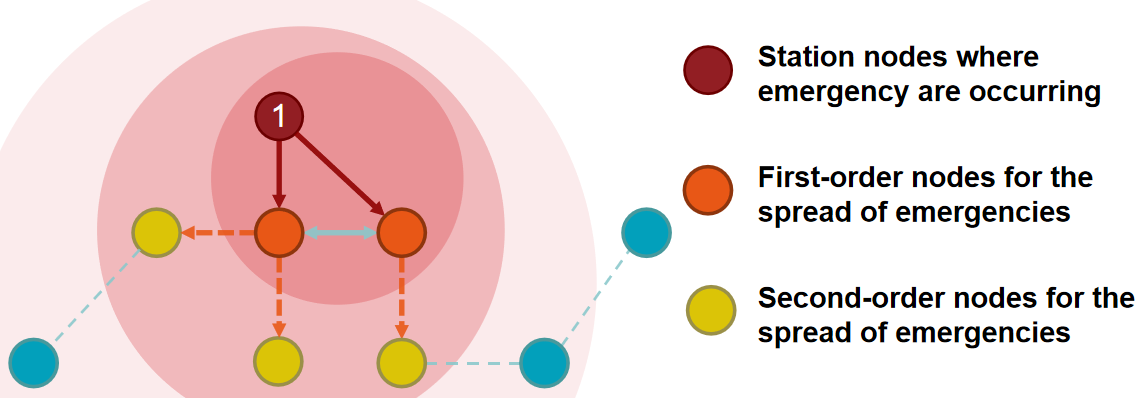}\\
	\caption{Diagram of the spread of an emergency event}
	\label{fig1}
\end{figure}

The emergencies studied in this paper often exhibit spatial diffusion, i.e., an emergent event in one node will spread to other nodes. We assume that the emergencies will spread along the edges in the emergency diffusion graph $G^\prime$. In the diffusion graph shown in Fig. \ref{fig1}, an emergency event occurs at node 1, which first affects the nodes directly connected to the orange nodes in the graph, followed by the set of nodes subconnected to yellow nodes, and then gradually spreads to the whole graph.

The characteristics of spatial dependence can be elaborated in terms of Euclidean spatial correlations and non-Euclidean correlations. Euclidean spatial distance is the straight-line distance between two notes, and Euclidean correlation means that if an emergency event occurs at a station node, it is more likely that an emergency event occurs in the area close to it. It is executed through the connectivity of Euclidean space during the operation process. For example, in an unmanned storage environment, the AGV might have a failure at the station node leaving it unable to move and operate normally, resulting in a timeout for the work of that node. It also affects the node's post-operational station node to  experience work timeout after a period of time. A non-Euclidean correlation means that the geospatial connection is not considered, but the two station nodes have other similarities, such as the station nodes have similar operational resources and use the same pipeline or line. Therefore, the two correlations can be described by the emergency event diffusion graph $G^\prime$.

b) Time dependence

Since an emergency event is a dynamic process that changes over time, it is typical of time-series data. The development of emergency event at the current moment is influenced by the situation at the previous moment and may also be correlated with data from earlier times. The development of emergent events is time-dependent.

\section{Human--Machine Collaborative Decision-making Algorithm for Dynamic Tasks}
In this section, we introduce the features for predicting the proliferation of emergencies in Section IV-A, the GACRNN-based key node prediction algorithm in Section IV-B, and the human--machine collaborative decision-making algorithm for dynamic tasks in Section IV-C.

\subsection{Characteristics of Emergency Event Diffusion}
In the problem of emergency disposal, identifying critical nodes and performing measures on them is important. We describe  critical nodes as two types of station nodes: nodes that are currently experiencing emergencies and nodes that will be affected by emergencies soon. Therefore, locating critical nodes is essentially a prediction of the state of node emergencies. The following describes the prediction features for predicting the emergency event state of node.

a) Resource feature $F_i^{r}$. The type and amount of resources of station nodes have an important effect on the development of node emergency events. Taking unmanned storage as an example, if a station node is storing explosive resources such as gasoline or hydrogen, a fire event might lead to the rapid spread of the emergency among other nodes. $F_i^{\mathrm{r\ }}$ consists of two attributes $\left(m_i,w_i\right)$ indicating the type of resources of the station node and the number of resources, in that order.

b) Central feature $F_i^c$. We have modeled the whole problem on a graph, with node importance reflecting the topology of the nodes and acting as a factor that affects the state of node emergencies. We use betweenness centrality, closeness centrality, and semi-local centrality for computing node importance. A node with high betweenness centrality is usually the key connection point in the graph and is calculated as shown in \eqref{eq2}, where $\sigma_{st}$ is the number of shortest paths from node $v_s$ to node $v_t$, and $\sigma_{s t}\left(v_{i}\right)$ represents the number of paths that pass through $v_i$ in the shortest path from $v_s$ to $v_t$ . Closeness centrality consists of in-closeness centrality and out-closeness centrality. The former is a measure of how easy it is for other nodes to get to a node by calculating the edges pointing to that node, with a higher closeness indicating how easy it is for other nodes to get to that point, as shown in \eqref{eq3}. The latter indicates how easy it is for a node to get to other nodes, as shown in \eqref{eq4}, $d_{ij}$ is the shortest path from the node $v_i$ to the node $v_j$, and $d_{ji}$ is the shortest path from $v_j$ to $v_i$. Semi-local centrality \cite{b24} considers the effect of the number of neighboring nodes and the clustering coefficient on the centrality of a node. It is calculated as shown in \eqref{eq5}, where $\Gamma(i)$ is the set of neighboring nodes of the node $v_i$; $\theta_{j}^{out}$ is the out-degree of $v_j$, which is the sum of the non-zero elements in the $j$ row of $A_{M\times M}$; and $f\left(c_{i}\right)$ is the function of the clustering coefficient $c_i$ of $v_i$. These three centrality calculations are collectively referred to as the centrality characteristics of station nodes.

\begin{equation}\label{eq2}
	C_{b}\left(v_{i}\right)=\sum_{s \neq v_{i} \neq t \in  V} \frac{\sigma_{s t}\left(v_{i}\right)}{\sigma_{s t}}
\end{equation}
\begin{equation}\label{eq3}
	C_{c-{out}}\left(v_{{i}}\right)=\frac{1}{\sum_{j} d_{ij}}
\end{equation}
\begin{equation}
	C_{c-{in}}\left(v_{{i}}\right)=\frac{1}{\sum_{j} d_{ji}}
	\label{eq4}
\end{equation}
\begin{equation}
	C_{r}\left(v_{i}\right)=f\left(c_{i}\right) \sum_{j \in \Gamma (i)}\left(\theta_{j}^{out}+1\right)
	\label{eq5}
\end{equation}

c) Task feature $F_i^\mathrm{w}$. The task feature is for a single station node, and the task progress of a node indirectly reflects the node's influence by the emergency events. Thus, the task failure rate of the node is taken as the task feature of the station node, with the calculation formula shown in \eqref{eq18}.

d) Event feature $F_i^{\mathrm{a}}$. The type of emergent event at the station node is one of the factors affecting the diffusion of emergent events. It consists of the node emergency event type. If the node does not have a contingency, $F_i^{a}=0$. In the case of an unmanned storage system, there are several event classifications: equipment failure, the failure of robots, conveyors, and other equipment in the unmanned storage system; insufficient resources such as boxes that lead to a delay in the shipping process; and fire events that can quickly spread and paralyze the whole operation. When an emergency event occurs at a station node, the type of emergency event at the node can be obtained through node sensors or staff reports.

We use the preceding four features to predict the state of the nodes' emergent events. $k_i$ indicates the state of the node's emergency event. The node's state is divided into three types ($k=0,1,2$): $k=0$ means that the node is currently normal; $k=1$ means that the node is currently having a burst; and $k=2$ means that the node's emergency event is over and the rescue fails, it also loses its diffusion capability.

\subsection{GACRNN-Based Key Node Prediction Algorithm}
Locating key nodes is conducive to better planning of emergency disposal plans to ensure that emergencies receive a response and resolution in the shortest possible time. As the diffusion of emergencies is time-dependent, we improve on the diffusion convolutional recurrent neural network (DCRNN) model \cite{b25} to obtain a new GAT-GCN-based recurrent neural network prediction (GACRNN) model and its set of key node predictions.

\begin{figure}[ht]
	\centering
	\includegraphics[width=9.1cm,height=5.05cm]{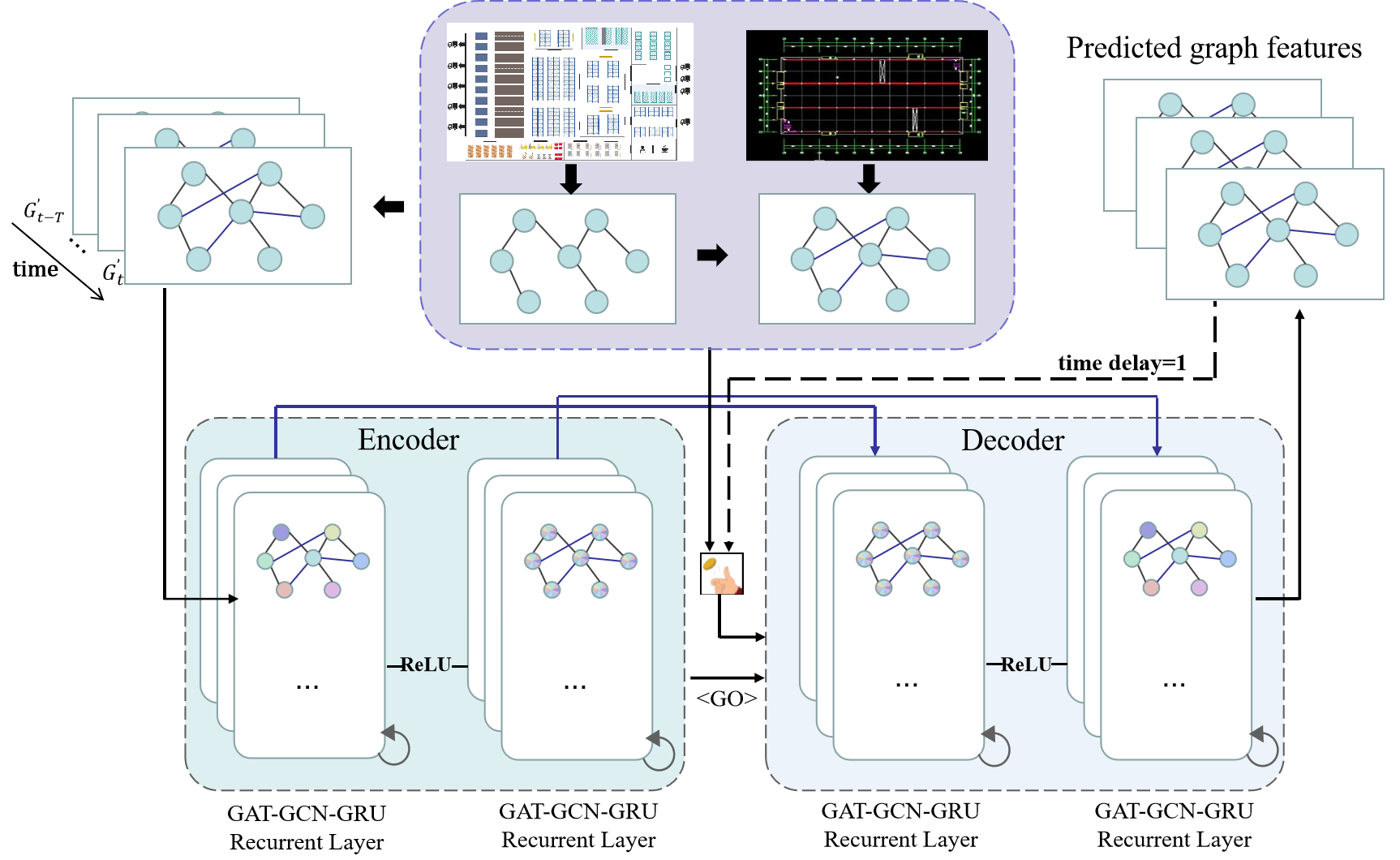}\\
	\caption{GACRNN model architecture diagram}
	\label{fig2}
\end{figure}

DCRNN is a neural network architecture for predicting time series data on graph-structured data and has been used with good experimental results in applications such as traffic flow and air quality prediction. DCRNN applies a combination of diffusion convolution and recurrent layers to the graph, with the diffusion convolution layer modeling spatial dependence and the recurrent neural network modeling temporal dependence so that it captures the spatial and temporal dependence between nodes. DCRNN introduces diffusion convolution to be able to capture non-Euclidean spatial dependencies, which require long training times. We model the spatial dependencies during the development of emergent events as a diffusion process among the station nodes on the graph, which includes non-Euclidean connections. Therefore, we improve DCRNN by introducing the GAT-GCN layer to replace the diffusion convolution layer to make it better suited to the key node prediction problem, thereby yielding the GACRNN model, as shown in Fig. \ref{fig2}.

After passing the GAT and updating the weights of the adjacency matrix, we  obtain the updated weight matrix $W_k$ and features $X_k$ of each node using $W_k$. After the GCN layer, the node eigenvalues are updated again, so the original graph convolution formula is transformed into

\begin{equation}
	g_{\mathrm{W}_{k^{*}}} \boldsymbol{X}_{k+1}=\tilde{\boldsymbol{D}}-\frac{1}{2} \tilde{\boldsymbol{A}} \tilde{\boldsymbol{D}}^{-\frac{1}{2}} \boldsymbol{X}_{k} W_{k}.
	\label{eq6}
\end{equation}

\begin{figure}[ht]
	\centering
	\includegraphics[width=9.0cm,height=3.3cm]{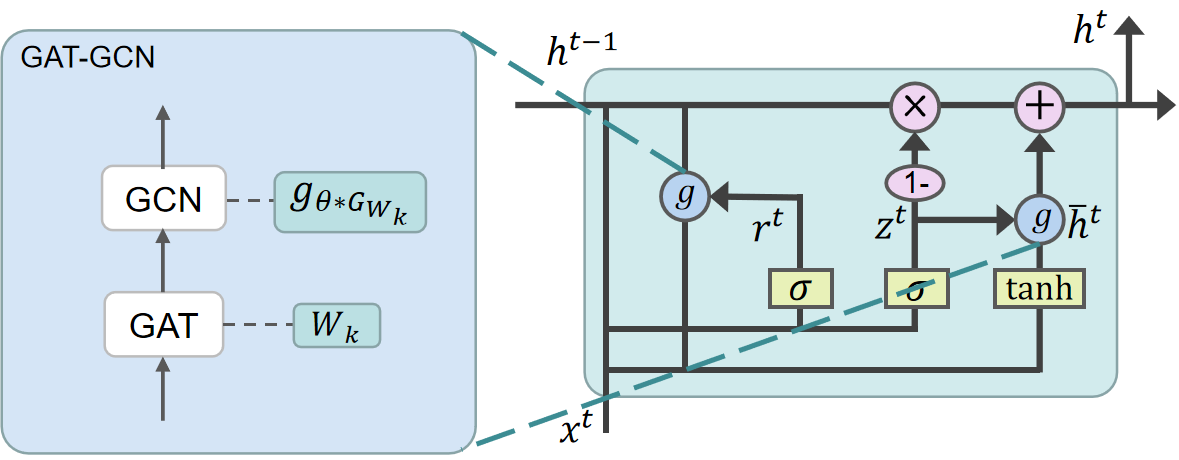}\\
	\caption{GAT-GCN-GRU structure diagram}
	\label{fig3}
\end{figure}

Using the feature mapping of the GAT-GCN layer as input, a network of gated recurrent units in a recurrent neural network captures the temporal dependencies between the timing data. The matrix multiplication in the GRU is replaced with the GAT-GCN to obtain a GAT-GCN-GRU based on the GAT-GCN-GRU, as shown in Fig. \ref{fig3}.

The reset gate and update gate equations for the GAT-GCN-GRU are as follows.

\begin{equation}
	r^{t}=\sigma\left(g_{\theta * G_{W_{k}}}\left[x^{t}, h^{t-1}+b_{r}\right]\right)
	\label{eq7}
\end{equation}
\begin{equation}
	z^{t}=\sigma\left(g_{\theta * G_{W_{k}}}\left[x^{t}, h^{t-1}+b_{u}\right]\right)
	\label{eq8}
\end{equation}

After the gating signal is obtained, the data is reset and then spliced with $x^t$, with the data then scaled by the activation function to obtain ${\bar{h}}^t$ as shown in \eqref{eq9}, where $\odot$ denotes the multiplication of the elements of the matrix. The final equation for the update process is shown in \eqref{eq10}.

\begin{equation}
	\bar{h}^{t}=\tanh \left(g_{\theta * G_{W_{k}}}\left[x^{t},\left(r^{t} \odot h^{t-1}\right)+b_{c}\right]\right)
	\label{eq9}
\end{equation}
\begin{equation}
	H^{t}=z^{t} \odot h^{t}+\left(1-z^{t}\right) \odot \bar{h}^{t-1}
	\label{eq10}
\end{equation}
The GAT-GCN convolution and sequence-to-sequence learning frameworks of GACRNN are used to capture the temporal and spatial dependencies of bursts to predict node states and thus obtain key nodes.

The emergency event diffusion graph $G^{\prime}=\left(V_{P}, E_{S}, D\right)$ and the graph signal $X\in R^{N\times I}$ consisting of the features are the inputs to DCRNN, where $I$ is the number of features of each node. According to the description in Section IV-A, $I=8$. $X_t$ denotes the graph signal at the moment $t$:
\begin{equation}
	X^{t}=\left[\begin{array}{cccccccc}m_{1} & w_{1} & c_{b 1} & c_{o 1} & c_{i 1} & c_{r 1} & F_{1}^{w} & F_{1}^{a} \\m_{2} & w_{2} & c_{b 2} & c_{o 2} & c_{i 2} & c_{r 2} & F_{2}^{w} & F_{2}^{a} \\\vdots & \vdots & \vdots & \vdots & \vdots & \vdots & \vdots & \vdots \\m_{N} & w_{N} & c_{b N} & c_{o N} & c_{i N} & c_{r N} & F_{N}^{w} & F_{N}^{a}\end{array}\right].
	\label{eq11}
\end{equation}

\begin{figure}[ht]
	\centering
	\includegraphics[width=8.6cm,height=6.98cm]{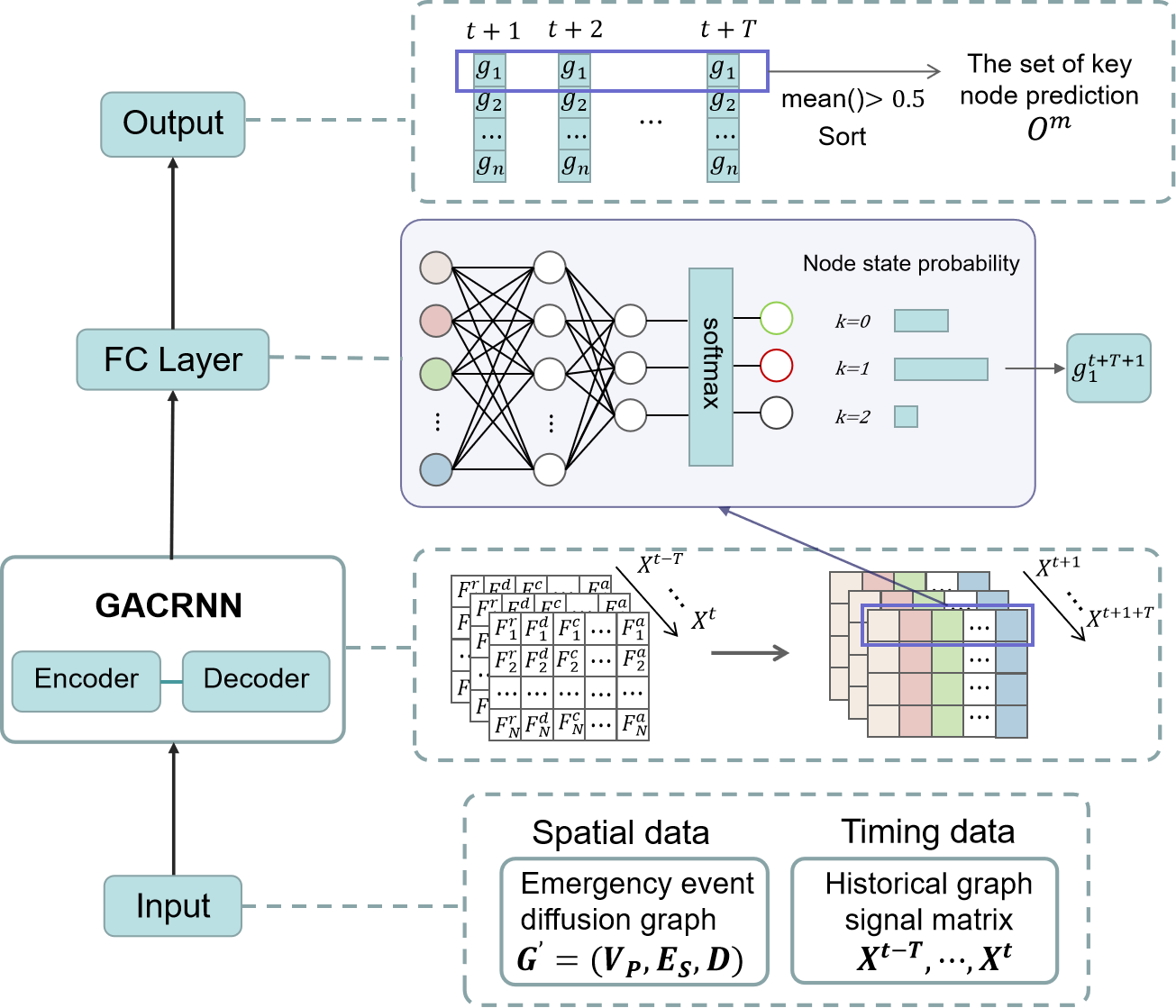}\\
	\caption{Key node forecasting framework diagram}
	\label{fig4}
\end{figure}

The predicted graph signal matrix is produced from the graph signal matrix and graph $G^\prime$ within the current time step. The predicted features of each node are connected to a fully connected layer with an output size of 3. The state of the emergency event at the next time of each node is predicted by the softmax function. Our method ranks the station nodes according to the probability $g>0.5$ that the predicted node is in state $k=1$ from largest to smallest and produces the set of key node predictions $O^m={(v_s,{\bar{g}}_s),\cdots,(v_t,{\bar{g}}_t)}$ at time $t$. The specific prediction framework is shown in Fig. \ref{fig4}.

\subsection{Human--Machine Collaborative Decision-Making Algorithm for Dynamic Tasks}\label{AA}
Since there will likely be errors in the set of key nodes predicted by the machine and uncertainty in the development of emergencies, task filtering is the basis of task scheduling. Furthermore, the task list must meet the requirements of the human decision-maker, which is essential for the human to trust the system and adopt the suggestions given by the intelligent system in emergency situations. To reduce the prediction error and determine the set of emergency disposal tasks, we introduce the idea of human--machine collaboration. Human decision-makers should be able to monitor the prediction results of the intelligent system in real time and provide modified feedback to the system by means of intervention.
The set of key node predictions $O^m$ leads to the set of emergency disposal task nodes $V^{mE}=\left\{v_s^\ast,\cdots,v_t^\ast\right\}$ predicted by the machine. We provide a human--machine interface that displays the machine-predicted key node list $V^{mE}$ to humans in real time, with the human decision-maker able to modify the list as shown in Fig. \ref{fig4}. Red entries indicate nodes experiencing an emergency event, and yellow entries indicate nodes predicted to experience an event.
\begin{figure}[ht]
	\centering
	\includegraphics[width=8.14cm,height=4.30cm]{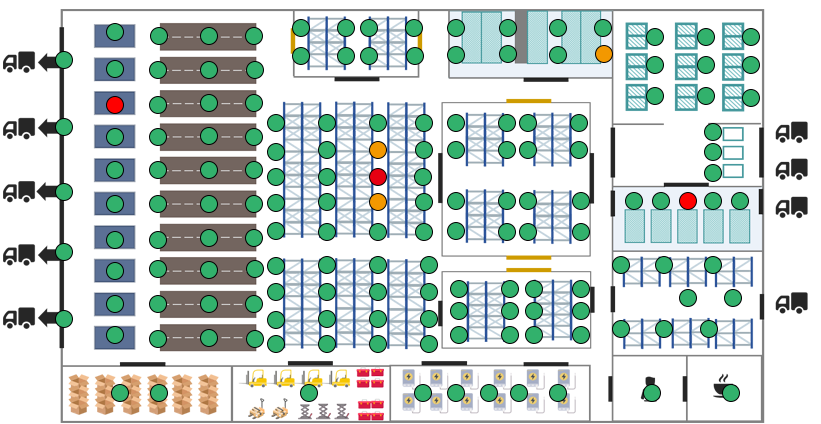}\\
	\caption{Human--computer interaction interface}
	\label{fig5}
\end{figure}
The interface has three basic operations for the human to consider as follows.
\begin{itemize}
	\item Add nodes: If the human determines the station node $v_i$ is considered important, but not in the set of affected nodes $V^{mE}$, the human may adjust the list of emergency disposal task nodes to include the additional nodes: $V^{hE}={v_s^\ast,\cdots,v_t^\ast,v_i^\ast}$.
	\item Delete node: If the human thinks the machine list includes a node $v_t^\ast$ corresponding to the station node ${v_t}$ that does not need to be disposed for the time being, the node can be deleted from $V^{mE}$ to yield $V^{hE}={v_s^\ast,\cdots,v_i^\ast}$.
	\item Reset list: If the human thinks that the disposal task list given by the machine does not match the current situation, the list can be reset. The model then predicts the key nodes based on current data.
\end{itemize}

The complete algorithm is shown in Algorithm 1. When using the model for predicting the emergency disposal task list, humans will correct the labels, so we improve the learning ability of the model by retraining the model offline. We also take into account that human feedback is sometimes wrong and that human corrections to labels are unqualified. Such human errors may lead to a cascade of errors and negatively affect the subsequent prediction if the label remains incorrect. The planned sampling in the DCRNN model solves this problem of weighing the human corrections and machine predictions.

\begin{algorithm}
	\caption{Human--machine collaborative decision-making algorithm for dynamic tasks.}
	\label{algorithm1}
	\begin{algorithmic}[1]
		\REQUIRE Resource feature $F_i^{\mathrm{r\ }}$; central feature $F_i^{\mathrm{c}}$; task feature $F_i^{\mathrm{w}}$; event feature $F_i^{\mathrm{a}}$; emergency event diffusion Graph $G^\prime$; the size of the task window set by the person $task\_size$. 
		\ENSURE the set of emergency disposal task $V^E$.
		\WHILE{$t<$ end time $T$}
		\STATE Obtain $F_i^{\mathrm{r}}$ and $F_i^{\mathrm{a}}$ from the station nodes in the scene. 
		\STATE Calculate $F_i^{\mathrm{c}}$ in the emergency disposal situation graph. 
		\STATE Calculate $F_i^{\mathrm{w}}$ based on \eqref{eq18}. 
		\STATE Pre-process the four features to obtain the graph signal matrix $X$.
		\STATE Invoke the GACRNN prediction model with input $\left[X^{(t-T+1)}, \cdots, X^{(t)} ; G^{\prime}\right] \rightarrow\left[\left(g_{1}, \cdots, g_{N}\right)^{(t+1)}, \cdots,\left(g_{1}, \cdots, g_{N}\right)^{(t+1+T)}\right]$.
		\STATE Sort by ${\bar{g}}_i$ from highest to lowest and intercept the top $task\_size$ as the set of key node predictions. 
		\STATE Obtain the emergency disposal task nodes $V^{mE}$ corresponding to the $O^m$
		\IF{the human decision maker modifies $V^{mE}$}
		\STATE  $V^E\gets V^{hE}$
		\STATE  Update the node states tag value
		\ELSE
		\STATE  $V^E\gets V^{mE}$
		\ENDIF
		\ENDWHILE
	\end{algorithmic}
\end{algorithm}

\section{Human Offline Experience-Constrained Emergency Disposal Task Scheduling Algorithm}

We now abstract the emergency disposal task scheduling problem as a decision problem of inserting emergency disposal tasks into the original work sequence and implement the solution in the framework of reinforcement learning. Using safety constraints with human offline settings performs safety action screening to achieve faster decision making. The problem modeling is introduced in Section V-A; the algorithm framework is introduced in Section V-B; the algorithm is introduced in Section V-C; and the overall human--machine collaborative emergency disposal method proposed in this paper is summarized in Section V-D.

\subsection{Problem Modeling}
We choose single-agent deep reinforcement learning to solve the emergency disposal task scheduling problem, with the agents as the decision makers for scheduling.

The process is modeled as a Markov stochastic decision process $(S,A,P(s^\prime\mid s,a),R,S^\prime)$, where $S$ denotes the set of states of the operational environment, $A$ denotes the set of actions of the insertable positions of the emergency disposal task nodes, $P(s^\prime\mid s,a)$ denotes the probability of moving to state $s$ after performing action $a$ in state $s^\prime$ probability (i.e., the action policy of the emergency disposal task node). $R$ denotes the reward function, which calculates the reward that the contingent task node receives by executing action $a$ in state $s$, i.e., the feedback from the operational environment to the emergency disposal task node for choosing to execute the action. $s^\prime$ denotes the next state updated by the operational environment after the emergency disposal task node chooses to insert the position action.

\textbf{Environment state.} From the definition of Markov decision, it is known that only the environment state at the current moment $t$ is considered in the decision, and the current environment state then contains the information of all previous states. Thus, the state of the station node to be inserted, the process subgraph, and the current position of the emergency disposal task node should all be considered in the scheduling of the emergency disposal task node. The state at time $t$ is represented by the triple $s^{t}=(V_{t}^{E}, K^{t}, m G^{t}, \Gamma^{t})$, where $V_{t}^{E}$, $K^t$, $mG^t$, and $\Gamma^{T}$ denote the task list, state of the station nodes, process subgraph information, and the current location information of the task node at time $t$, respectively. \eqref{eq13} denotes the emergency disposal task list at time $t$, and \eqref{eq14} denotes the state of the station node. \eqref{eq15} denotes the process subgraph corresponding to the station node at time $t$, and $mg_i=(nw_i,e_i^{mg})$, where $nw_i$ and $e_i^{mg}$ reflect the tasks of the station node at this time, and the position where the emergency disposal task can be inserted, respectively. \eqref{eq16} denotes the current position of the emergency disposal task node, where $eV_i\in V^E$ denotes the list of emergency disposal task nodes. $in\_e$ is the set of nodes being pointed to the node. $out\_e$ is the set of nodes pointing to the task node. $in\_e=out\_e=\emptyset$, if the current emergency disposal task node is not inserted into the corresponding process subgraph. At moment $t$, environment state $s^t\in S$ is given by \eqref{eq12} through \eqref{eq16}.
\begin{equation}
	s^{t}=\left(V_{t}^{E}, K^{t}, m G^{t}, \Gamma^{t}\right)
	\label{eq12}
\end{equation}
\begin{equation}
	V_{t}^{E}=\left(e V_{0}, \cdots, e V_{n}\right)
	\label{eq13}
\end{equation}
\begin{equation}
	K^{t}=\left\{k_{1}, \cdots, k_{N}\right\}
	\label{eq14}
\end{equation}
\begin{equation}
	mG^{t}=\left\{mg_{i}, \cdots, mg_{j}\right\}
	\label{eq15}
\end{equation}
\begin{equation}
	\Gamma^{t}=\left\{{}^{\prime}V_{0}^{*\prime}:[{in\_e,out\_e}], \cdots,{}^{\prime} V_{n}^{* \prime}:[{in\_e},{out\_e}]\right\}
	\label{eq16}
\end{equation}

\textbf{Action space.} When the agent is in state $s^t$, it needs to execute the scheduling decision action by inserting the node in the emergency disposal task list into the appropriate position in the process subgraph and then obtaining the tasks and works sequence for each station node. To avoid breaking the planned work order, we use edges in the subgraph to represent insertable positions. At time $t$, the optional location set $P_i$ of the emergency disposal task node $v_i^\ast$ depends on process subgraph $mg_i$, namely $|P_i|=|e_i^{mg}|+2$. (If $mg_i=\emptyset$, then $|P_i|=1$.) Therefore, at time $t$, the size of the action space changes dynamically with the progress of the operation and the changes in the operation environment. The action space $a^t\in A$ is expressed as 
\begin{equation}
	a^{t}=\left\{\left(e V_{0}, p_{0}\right), \cdots,\left(e V_{n}, p_{n}\right)\right\}, e V_{n} \in V^{E}.
	\label{eq17}
\end{equation}

\textbf{The reward function.} Reinforcement learning is the process of learning through continuous ``trial and error'' by interacting with the environment and learning from the rewards received by the agent’s actions. Thus, designing a suitable reward function is an important part of reinforcement learning, with the reward function meeting the optimization objectives of the problem. The emergency disposal problem has two objectives: a) minimize the failure of the originally planned tasks and b)  minimize the effect of the emergency events.

The objective function in this paper consists of three parts: the normal work progress, the effect of the emergency events, and the normal node resource situation. We describe these in the following paragraphs. 

Normal work progress $R_{work}$: the work completion rate $p_i^{succ}$ of each station node is used to measure the progress of normal works. Station nodes can be classified into three categories according to each node's state. States $k=0,1$ indicate normal work arrangements, so the work completion rate of station node $v_i$ is calculated as shown below, where ${sv}_i$ denotes the job nodes belonging to node $v_i$ in the set of completed work nodes:

\begin{equation}
	p_{i}^{succ}=\left\{\begin{array}{cc}\frac{\left|s v_{i}\right|}{\left|n w_{i}\right|+\left|s v_{i}\right|}, & k_{i} \leq 1 \\0, & k_{i}=2\end{array}\right .
	\label{eq18}
\end{equation}

Since not all station nodes have work assigned, we use the average work completion rate for station nodes with process subgraphs to express the final normal work progress reward value, calculated as 

\begin{equation}
	R_{work}=\frac{{\sum_{i=1}^{N} p_{i}^{succ}}|_{mg_{i} \in mG}}{|mG|}.
	\label{eq19}
\end{equation}

The effect of emergency events ${R}_{e}$: The effect of an emergency event can be expressed by the information in the graph. A node with $k=1$ has not yet completely damaged, but the corresponding process subgraph already has failed work nodes. The failed task node is isolated and has a negative value, and this negative value can be considered the negative impact caused by the emergency event. When an emergency event occurs at a station node, the total value cost of the node and its task nodes decreases over time without the intervention of a disposal action. When the state of a station node progresses to $k=2$, the node loses its own value and at the same time loses its connectivity role in the physical space and on the pipeline. The connectivity role is proportional to the node's own resources and inversely proportional to the distance on the edge, so the value $w_{ij}$ can be calculated using \eqref{eq20}.
\begin{equation}
	w_{i j}=w_{i} \cdot \frac{1}{d_{ij}}
	\label{eq20}
\end{equation}
\begin{equation}
	\left.w_{i}\right|_{k_{i}=3}=-\sum_{j=1}^{N}\left(w_{ij}+w_{ji}\right)
	\label{eq21}
\end{equation}
In the preceding equations, $w_i$ denotes the resources value of node $v_i$, and $d_{ij}$ denotes the distance from node $v_i$ to $v_j$ when the two nodes are directly connected. When $k_i=2$, ${cost}_i<0$. $R_e$ is calculated as shown in \eqref{eq22}.

\begin{equation}
	R_{e}=\sum_{i=1}^{N}\left(\left.{cost}_{i}\right|_{v_{i} \in V^{a}}\right), N=\left|V^{P}\right|
	\label{eq22}
\end{equation}
\begin{equation}
	{cost}_{i}=w_{i}+\sum_{j=1}^{n u m} w_{i j},{num}=\left|n w_{i}\right|
	\label{eq23}
\end{equation}

Normal node resource situation ${R}_{n}$: This value describes the resource situation of normal station nodes. If $V^s$ denotes the set of station nodes not affected by the emergency events, i.e., $V^s=V^P-{v_i\mid v_i\in V^a\parallel et_{in}=0}$, then $R_n$ is calculated as shown in \eqref{eq24}.
\begin{equation}
	R_{n}=\sum_{i=1}^{N}\left(\left.w_{i}\right|_{v_{i} \in V^{s}}\right), N=\left|V^{P}\right|
	\label{eq24}
\end{equation}

There are trade-offs between maintaining the original operational processes, suppressing the negative effect of emergencies, and protecting the station nodes. We should prevent biasing decision making of the agent toward one of the reward methods and also consider the needs of the actual scenario. For example, in the case of unmanned storage, when a large, rapidly developing event such as a fire occurs, we should abandon (de-prioritize) the importance of maintaining the original task and focus on limiting the scope of the fire. In the case of a smaller congestion event, normal work should continue as much as possible with less importance attached to the emergency. Therefore, we introduce the weight coefficient $\beta$ to enable the relative relationship of the three components in the objective function adaptively updateable for different environmental states \cite{b26}, as shown in \eqref{eq25}. The human can also set the weight coefficient through the human--computer interaction interface.

\begin{equation}
	\beta=\sigma\left[R_{work}^{t}-\left(R_{n}^{t}+R_{e}^{t}\right)\right], \beta \in[0,1]
	\label{eq25}
\end{equation}

In \eqref{eq25}, $\sigma$ is a sigmoid function. The formula for combining the three components is shown in \eqref{eq26},

\begin{equation}
	Z^{t}=\beta R_{\text {work }}^{t}+(1-\beta) \eta\left(R_{n}^{t}+R_{e}^{t}\right),
	\label{eq26}
\end{equation}
where $\eta$ is the normalization operation. The cumulative reward reflects the long-term effect, which is the objective of RL maximization. Thus, the reward function is defined as
\begin{equation}
	r^{t}=Z^{t+1}-Z^{t}.
	\label{eq27}
\end{equation}

\subsection{Human--Machine Collaboration of Emergency Disposal Task Scheduling Algorithm Framework}

By the previous definition of modeling the emergency disposal task scheduling problem, the problem is transformed into a problem of inserting emergency nodes into the original task sequence, i.e., a sequential decision problem. Deep reinforcement learning has a self-learning capability to adapt to changing environments in real time, dynamically learning strategies from the feedback provided by the environment, and handles large-scale optimization problems, making it very effective for problems such as emergency disposal that require timely processing of large amounts of data. Therefore, the Double DQN algorithm is chosen to solve the problem of scheduling for emergency disposal and adaptation to the dynamic changes in the environment.

\begin{figure}[ht]
	\centering
	\includegraphics[width=8.66cm,height=5.40cm]{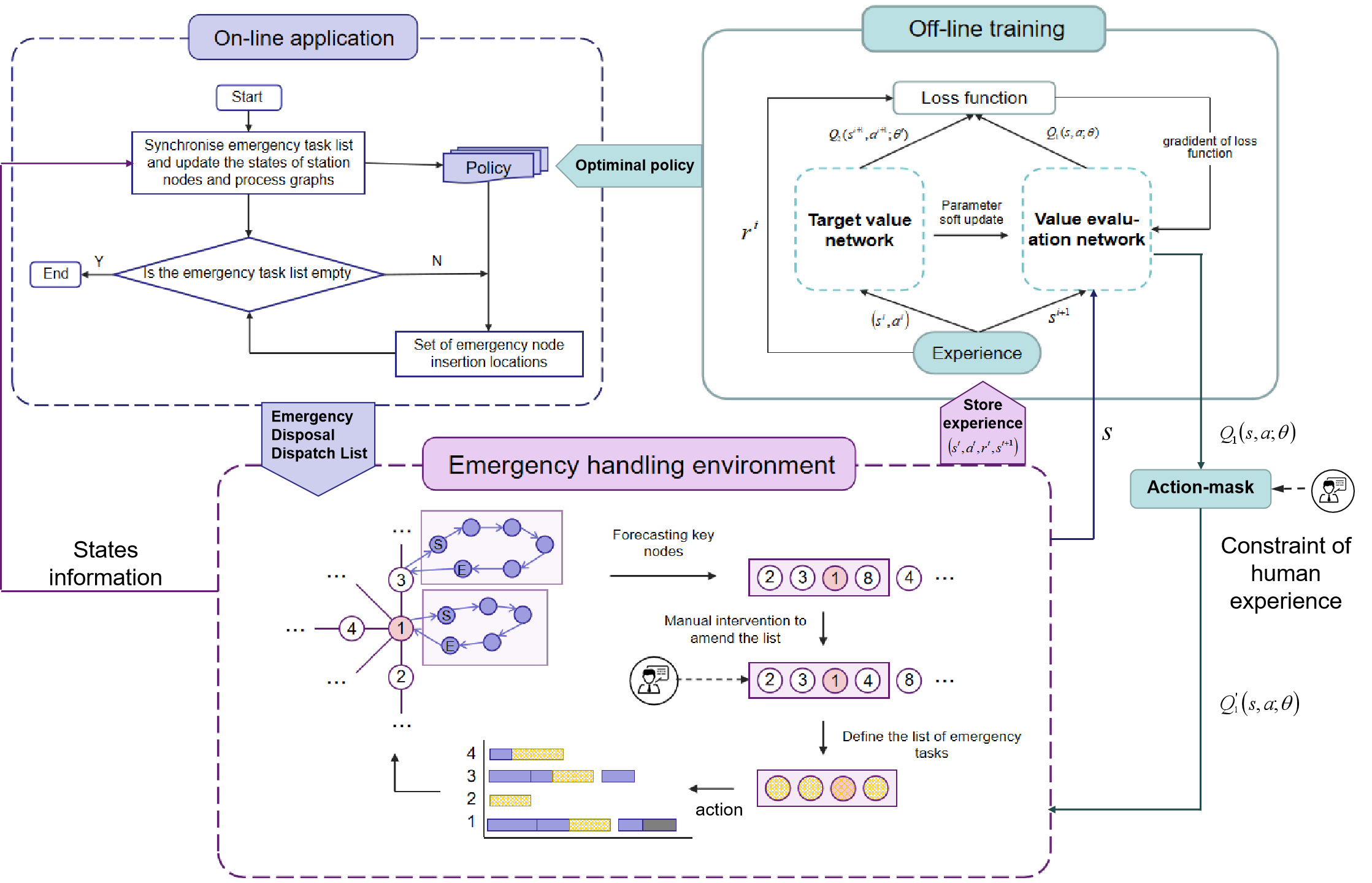}\\
	\caption{An algorithmic framework for scheduling emergency disposal tasks with human experience constraints}
	\label{fig6}
\end{figure}

However, deep reinforcement learning algorithms require a long training time, and the training is conducted by trial and error in the early stages, which is unsuitable for direct use in emergency response. In this paper, we propose an ``offline learning-online use'' framework to make deep reinforcement learning algorithms more suitable for emergency response problems. The framework includes three parts: scheduling environment, offline learning, and online application, as shown in Fig. \ref{fig6}.

Although offline learning requires a long training time, it is able to learn the optimal policy and apply it to new scheduling problems. In the offline learning phase, an offline training strategy is performed for the double DQN-based emergency disposal task scheduling algorithm. In the online use phase, the state information is updated according to the information in the situation, and the optimal scheduling policy is used to make online decisions on the emergency disposal task list obtained from the task allocation process to obtain the optimal solution.

\subsection{Human Experience-Constrained Emergency Disposal Task Scheduling Algorithm}

In the process of emergency disposal, there are often strict constraints relating to safety, technical, and feasibility concerns, which we refer to collectively as human experience constraints. Using unmanned storage as an example, when the AGV is performing the process of taking out flammable goods, placing them on pallets, and encapsulating them, if an emergency event occurs near the node with the possibility of spreading, the process should consider the flammability of the goods, encapsulate them, and then perform the emergency disposal task. In the case of a conveyor belt failure that prevents normal operation, scheduling the disposal task after normal operating hours means delaying normal operations that would eventually affect work at other stations. The safety  and feasibility constraints avoid these actions. However, the Double DQN algorithm through the neural network output action Q-value set is unable to consider whether the current action is safe and can be executed. The agent may choose an unsafe action, reducing the efficiency of emergency disposal, or even leading to further expansion of the emergency. It can also affect the learning efficiency of reinforcement learning, leading to a long convergence time of the neural network.

Therefore, the Action-Mask mechanism \cite{b27} is designed to process the action space, where each insertion position before the agent selects an action determines its safety and feasibility. If it does not meet the human experience constraint, then the action cannot be selected, and its Q value is assigned to the smallest value of negative infinity to prevent it from being selected. The human experience constraint is based on the safety judgment condition summarized offline by human experience, and the mask is added to the action that fails to pass the constraint. The Action-Mask mechanism not only ensures the safety of agent’s action but also screens the effective action space to shorten the time of emergency disposal decision and meet the requirements of emergency disposal safety and urgency.

\begin{figure}[ht]
	\centering
	\includegraphics[width=8.75cm,height=6.07cm]{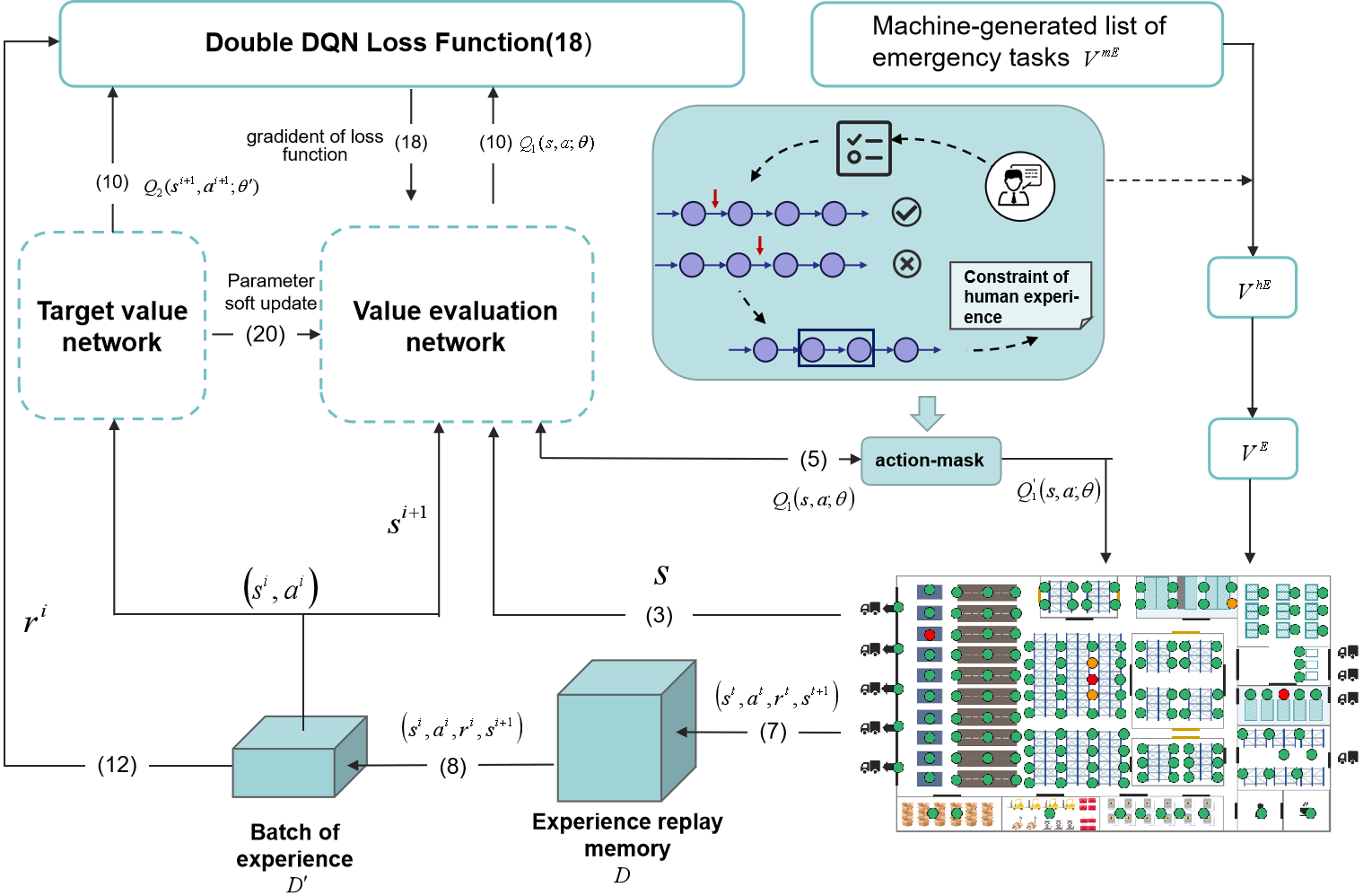}\\
	\caption{Flow chart of the Double DQN-based emergency disposal task scheduling algorithm}
	\label{fig7}
\end{figure}

The flow of the emergency disposal task scheduling algorithm is shown in Fig. \ref{fig7}, with the numbers on the connections in the figure referring to line numbers in Algorithm 2.

We now describe the emergency disposal task scheduling algorithm based on Double DQN. In the exploration phase of the emergency disposal task scheduling algorithm, the agent selects the insertion position of the emergency disposal task node as the execution action according to the node state and the process subgraph environment, with the specific operation being to input the current environment state into the neural network and obtain the output of the network. The goal is to obtain values corresponding to all actions. The action selection is first processed by the Action-Mask mechanism for the action space and then selected and executed using an $\varepsilon$-greedy strategy. 

The execution of specific actions is divided into three cases: a) if the process subgraph of node is empty or the node is having an emergency event, the emergency disposal task node is inserted directly; b) if the emergency node is already in the process subgraph, the agent adjusts its position; c) if the process subgraph of the node is not empty and the emergency node is not in it, the agent selects the position for insertion. The emergency node then executes the action selected by the agent. Then, the environment will update the node and process subgraph states. The reward model evaluates the reward value generated by the executed action, as in lines (5-6). Finally, the state transitions are recorded into the experience return memory for training the value evaluation network, which uses the values $Q$ obtained from the value evaluation network and the values ${Q}^\prime$ obtained from the target value evaluation network and their rewards to calculate the loss, updating the network parameters using the  gradient descent of the value evaluation network through the loss, as in lines (7-20).
\begin{algorithm}
	\caption{Double DQN-based dynamic scheduling algorithm for emergency disposal.}
	\label{algorithm2}
	\begin{algorithmic}[1]
		\REQUIRE Experience replay memory capacity $C$, reward discount factor $\gamma $, exploration rate $\varepsilon$. 
		\STATE Initialization: empirical experience replay memory $D$ of capacity $C$, initialized with random parameter sets $\theta$ and ${\theta}'$.
		\FOR{episode=1 TO $M$}
		\STATE Initialize the current state and emergency state and get the emergency disposal task nodes.
		\WHILE{timesep $t$ in $T$} 
		\STATE Mask the $Q$ value of an unsafe action via the Action-Mask mechanism and use the $\varepsilon$-greedy strategy to select an action $a^t$ to explore from the action space after the Action-Mask mechanism has masked the value. 
		\STATE Calculate the immediate reward $r^t$ based on \eqref{eq27} and observe the new state $s^{t+1}$. 
		\STATE Perform the storage state transition $(s^t,a^t,r^t,s^{t+1})$ to experience replay memory $D$.
		\STATE Select a random batch of training samples $D^\prime$ from $D$.
		\WHILE{$(s^i,a^i,r^i,s^{i+1})$ in $D^\prime$} 
		\IF{$s^{i+1}$ is end state}
		\STATE  $y^i=r^i$
		\ELSE
		\STATE   Choose the larger value of $Q_1$ and $Q_2$ as the action $a^{i+1}=\arg \max \left(Q_{1}\left(s^{i+1},a;\theta\right)+Q_{2}\left(s^{i+1},a;\theta^{\prime}\right)\right)$. 
		\STATE   Calculate target values based on target value evaluation networks $y^{i}=r^{i}+\gamma Q_{2}\left(s^{i+1}, a^{i+1} ; \theta^{\prime}\right)$.
		\ENDIF
		\STATE Perform the gradient descent on the loss function $Loss={(y^i-Q(s^i,a^i;\theta))}^2$ and update the value evaluation network parameters by back-propagating the gradient direction of the neural network $\theta^\prime$.
		\ENDWHILE
		\STATE After every $L$ decision steps, update the target value network parameter $\theta$ with the value network parameter $\theta^\prime$, i.e., $\theta^\prime\gets\theta$
		\ENDWHILE
		\ENDFOR
	\end{algorithmic}
\end{algorithm}
After a specified number of training sessions, the neural network model is able to obtain an excellent decision-making agent, which can be applied to the dynamic scheduling problem of emergency disposal of emergencies with good results. In this model, the agent makes decisions for the insertion position of the emergency disposal task nodes in the node corresponding to the process subgraph in the task allocation list.

\subsection{Human--Machine Collaborative Emergency Disposal Method}

The emergency disposal of emergency events studied in this paper is based on the task filtering by connecting the emergency disposal task nodes in the appropriate process subgraph in the emergency disposal situation graph to control the development of emergency events and solve them while trying to ensure the completion of normal work steps.

\begin{figure}[ht]
	\centering
	\includegraphics[width=8.68cm,height=5.18cm]{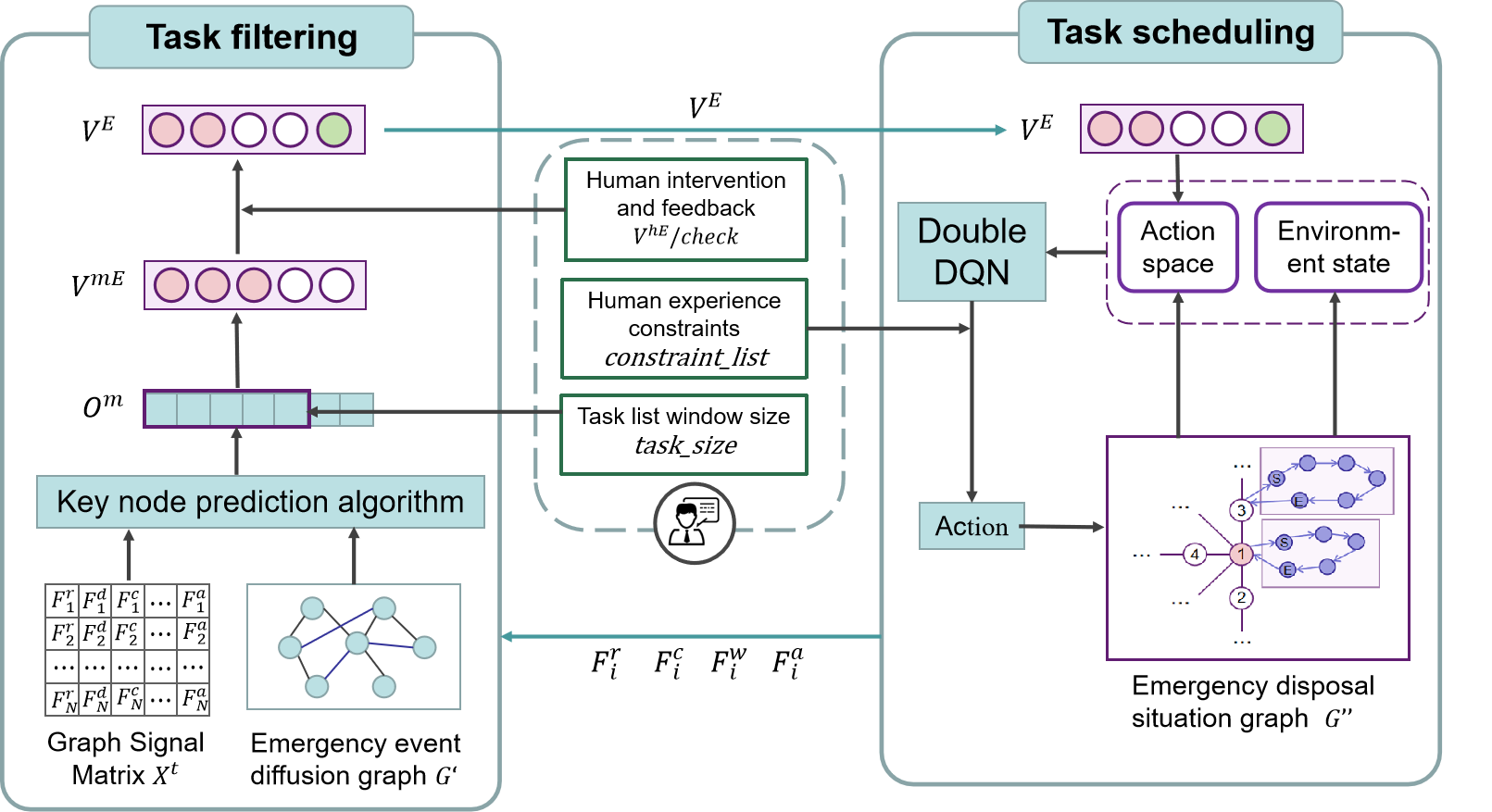}\\
	\caption{human--machine collaboration for emergency disposal}
	\label{fig8}
\end{figure}

The process of resolving emergencies is divided into two phases: task filtering and task scheduling. The method of the two phases is shown in Fig. \ref{fig8}. The task filtering phase assigns emergency disposal tasks by predicting key nodes, and the list of emergency disposal task nodes $V^E$ is used as an input to the emergency disposal algorithm. The station node resource features $F_i^{r}$, center features $F_i^{c}$, and station node task features $F_i^{w}$ calculated in the emergency disposal situation graph are sent to the GACRNN model prediction for the next round of key node prediction. 

\section{Experiments}

 To evaluate our proposed algorithm, we conducted experiments using an unmanned storage facility as the test situation. In this section, we introduce the unmanned storage simulation experimental scenario in Section VI-A; test and analyze the results of the algorithm in the task filtering phase in Section VI-B; test and analyze the results of the algorithm in the task scheduling phase in Section VI-C; and verify the role of man--machine coordination and the task filtering phase using ablation experiments in Section VI-D.

\subsection{Introduction of the Situation and Experimental Setup}

An unmanned warehouse refers to a warehouse setup built using automation, robotics, and artificial intelligence to perform operations such as storage, picking, transportation, and sorting of goods without human labor. However, due to factors such as a large storage area, large number of goods, concentrated storage of dangerous goods, and a tight workflow arrangement, such warehouses are prone to emergency events such as machine failure, changes in operational priority, and small accidents. Timely emergency disposal is needed to deal with emergencies to ensure the safety of goods and the normal operation of the work process.

\begin{figure}[ht]
	\centering
	\includegraphics[width=7.13cm,height=4.61cm]{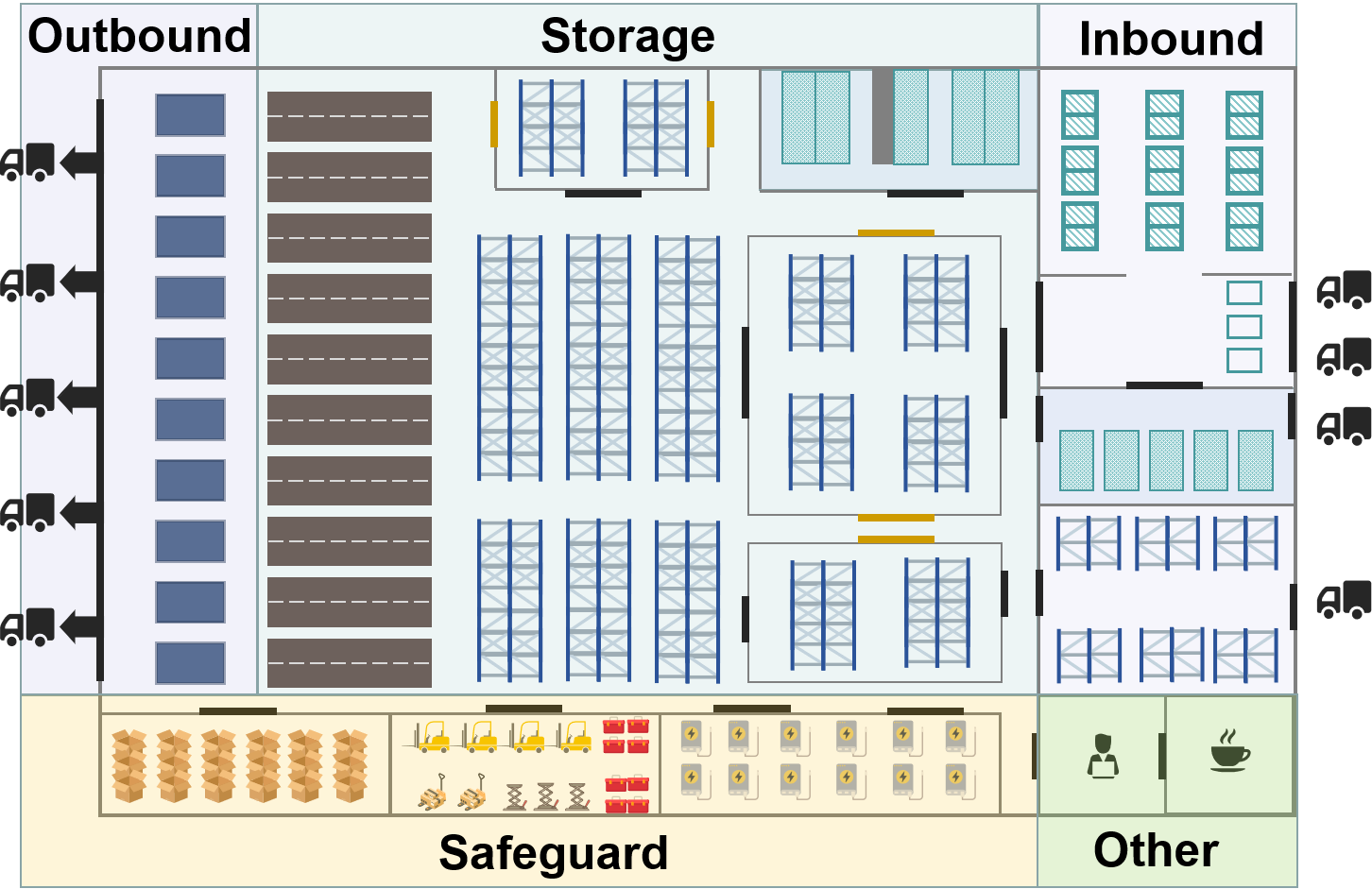}\\
	\caption{Storage floor plan}
	\label{fig9}
\end{figure}

The work areas of an unmanned warehouse are composed of storage, outbound, safeguard, and other areas. Each station in each area has a corresponding operational process. As each operation process has strict time and sequence requirements, emergency events can affect the normal work. The warehousing plan is shown in Fig. \ref{fig9}.

\begin{figure}[ht]
	\centering
	\includegraphics[width=7.76cm,height=3.68cm]{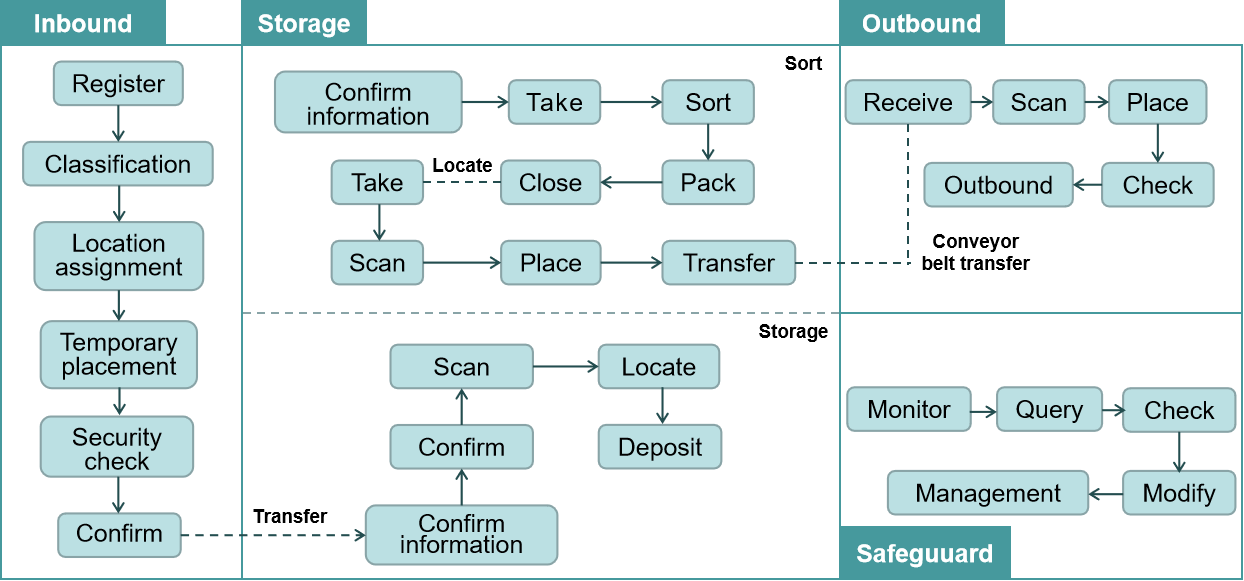}\\
	\caption{Unmanned storage area operational process diagram}
	\label{fig10}
\end{figure}

In the experimental warehouse, there are 150 station nodes, including 25 nodes in the inbound area, 80 nodes in the storage area (including 70 storage nodes and 10 picking nodes), 25 nodes in the outbound area, 8 nodes in the security area, 2 nodes in other areas, and 10 road mobile nodes. The operational process arrangement of each area is shown in Fig. \ref{fig10}. The road mobile and conveyor stations have only one mobile operation node, and the other areas are mainly rest areas without an operational process arrangement.

In the unmanned warehouse simulation, we analyze the actual situation and design three common types of emergency events: a) packing material shortages, which are material shortage contingencies at the picking and selecting stations in the storage area; b) equipment failures, which refer to the failure of equipment (such as shelves, conveyors, and scanners) or AGV vehicles performing work in the unmanned warehouse; and c) cargo fires, which are contingencies that occur in the storage area, and since fire develops faster than other events, the loss of node resources increases with time.

\subsection{Testing and Result Analysis of the Algorithm in the Task Filtering Phase}

The critical node prediction model of the emergency disposal task assignment phase was trained using the dataset simulated in the mockup platform. The dataset contained four dimensions of features such as node resource features, node vulnerability features, and center features, as well as station node location and connectivity information. Due to the time urgency of emergency disposal, 3 seconds was set as a time step in the simulation, i.e., each minute contained information for 20 time steps. The feature data of the next minute was predicted using the data of the past minute, i.e., the size of time step and prediction time step was 20. The data was divided into training and test sets in an 8:2 ratio. Min-Max normalization was applied to them to normalize the data and speed up the convergence of the model.

To test the GACRNN key node prediction algorithm  comprehensively, four metrics (Accuracy, Precision, Recall, and F1-Score) were used to measure the prediction results of node states.

\begin{figure}[ht]
	\centering
	\includegraphics[width=6.54cm,height=4.94cm]{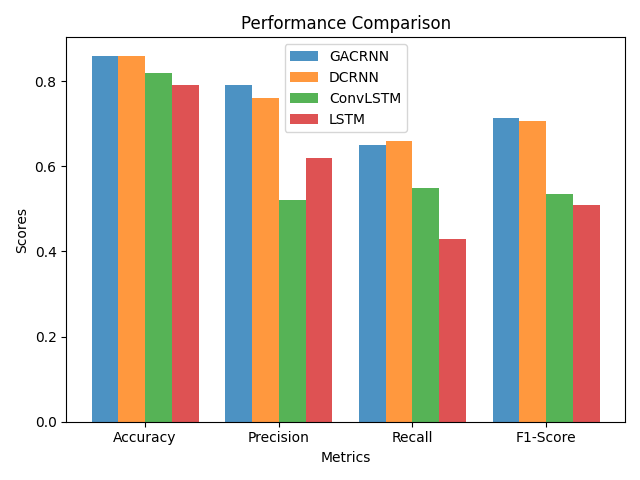}\\
	\caption{Experimental results of GACRNN and three benchmark models}
	\label{fig11}
\end{figure}

To compare and analyze the GACRNN node state prediction model, three benchmark models were selected for comparison, including DCRNN, ConvLSTM \cite{b28}, and LSTM. The benchmark models are all multi-classification models to predict the state of nodes. The test results are shown in Fig. \ref{fig11}, with detailed result data shown in Table \ref{tab-1}.

The figure shows that the model in this paper did not differ much from the three benchmark models in terms of accuracy, with all of them reaching more than 78\%, and GACRNN and DCRNN notably reaching 86\%. However, the process revealed that because DCRNN uses diffusion convolution, it extracted spatial dependencies better but with a longer training time. While GACRNN, DCRNN, and ConvLSTM (three models capable of spatio-temporal prediction) had better overall results, while LSTM only obtained temporal features and had slightly worse overall performance. Our GACRNN performed better in all aspects and was able to predict states relatively well. 

\begin{table}[htbp]
	\centering
	\caption{Detailed experimental results of GACRNN compared with the benchmark models}
	\begin{tabular}{lrrrr}
		\toprule
		Model & \multicolumn{1}{l}{Accuracy} & \multicolumn{1}{l}{Precison} & \multicolumn{1}{l}{Recall} & \multicolumn{1}{l}{F1-Score} \\
		\midrule
		GACRNN & \textbf{0.86} & \textbf{0.79} & 0.65  & \textbf{0.713} \\
		DCRNN & \textbf{0.86} & 0.76  & \textbf{0.66} & 0.706 \\
		ConvLSTM & 0.82  & 0.52  & 0.55  & 0.535 \\
		LSTM  & 0.78  & 0.62  & 0.43  & 0.508 \\
		\bottomrule
	\end{tabular}%
	\label{tab-1}%
\end{table}%

\subsection{Testing of Algorithms and Analysis of Results in the Task Scheduling Phase}

To evaluate the performance of the human experience-constrained Double DQN algorithms (HEC-DDQN) with security constraints we have proposed, we performed training and result analysis in a simulation scenario. During model training, one task cycle was set to 600 time steps (30 minutes). In the initial state, work steps were arranged for each region according to the work schedule, and stations were randomly selected within the region. Three emergency event sizes were set for the test: an initial 4 events with no misfire events called Scale 1, 4 events with misfire events called Scale 2, and 6 events with no misfire events called Scale 3. The corresponding settings are shown in Table 5.3. The human experience constraint was set to not perform the emergency disposal task of equipment failure before the shutdown process and to prioritize the emergency disposal task when the equipment failure occurred at the conveyor station note.

We tested our implementation of simulation experiments using Python 3.7, Gym, and the Double DQN algorithm using Keras, Tensorflow, and other frameworks to determine which offered the best performance. We used Tensorboard to plot the experimental results. The parameter setting part of the experiment in DRL followed the general principle, and the parameter settings with the best results were selected for multiple experiments. The maximum capacity of the experience replay memory was set to $C=2000$, the number of small batch samples was 32, and the total number of training cycles was 1500. Since we wanted the maximum final total reward, the discount factor was set as small as possible, to $\gamma$=0.97. To allow the model to explore the strategy fully in the initial stage, $\varepsilon$=0.9 was set as the initial value. In the subsequent training phase, the agent gradually learned better strategies and explored less, so the exploration rate was set to decay to 0.1 with a discount rate of 0.9995 during the training process.

(1) Analysis of algorithm convergence

\begin{figure}[ht]
	\centering
	\includegraphics[width=8.43cm,height=5.24cm]{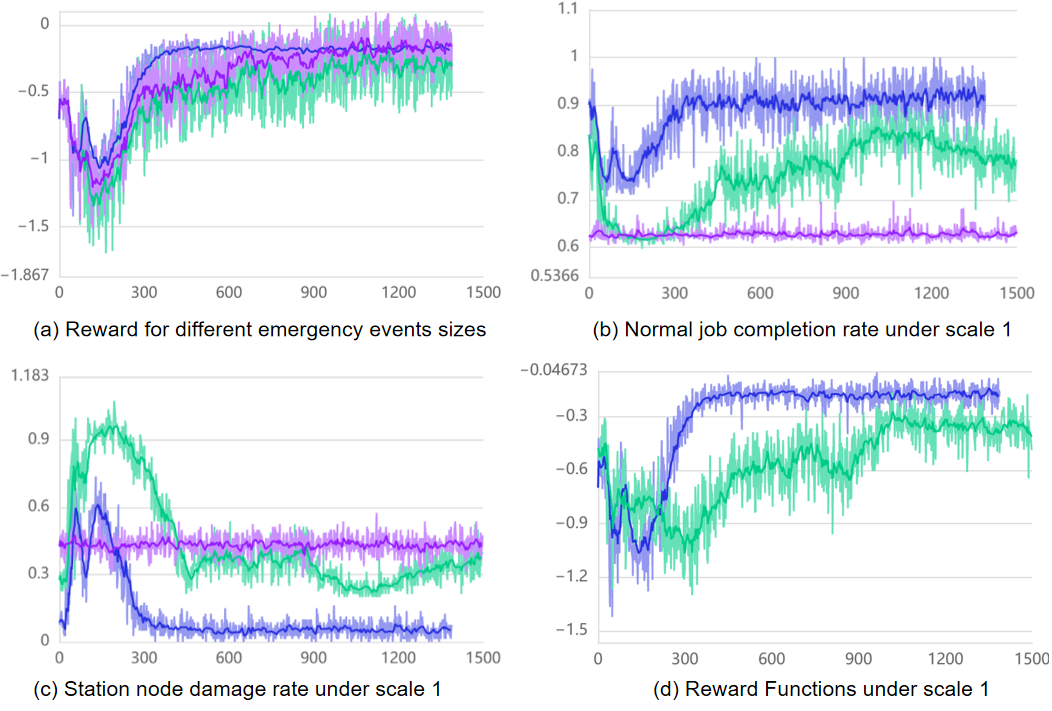}\\
	\caption{Experimental results on algorithm convergence and emergency disposal efficiency}
	\label{fig12}
\end{figure}

In this section, different numbers and types of emergency events were represented by different scales and were learned using the model in this paper. The reward curves obtained from the simulation tests are shown in Fig. \ref{fig12}(a), with the curve of the change in reward function for scale 4 in blue, the curve for scale 2 in green, and the curve for scale 3 in purple. The curves fluctuated under the three scales of events, but the overall increase tended to be stable. As reinforcement learning is learned through `exploration by trial and error,' the reward curve is not smooth. Therefore, we conclude that our model converges approximately for all sizes of emergency events.

The figure shows that in the case of small-scale emergency events, the increase in the reward function was more significant, but as the scale of the event increased, the growth of the reward function became slower. For the scenario with a fast spread of the emergency event scale 2, the reward function growth was the slowest among the three scenarios and had the lowest reward value. Our model performed better in emergency disposal for small-scale emergencies with slow development.

(2) Analysis of the effectiveness of emergency disposal

We also evaluated the emergency disposal effect of the emergency event using the completion rate of Normal work $R_{work}$, the failure rate of nodes ${rate}_{fail}$, and the reward function $r$ as evaluation indicators. The calculation formula of the reward function and the completion rate can be found in formula \eqref{eq27} and \eqref{eq19}, respectively. The calculation formula for the failure rate of station nodes is shown below.

\begin{equation}
	{rate}_{fail}=\frac{\left|V^{a}\right|}{\left|V^{E}\right| \times t_{\max}}
	\label{eq28}
\end{equation}

To evaluate the performance of our proposed scheduling method in terms of the effectiveness of emergency disposal, the task filtering model obtained after training in the previous section was used to generate the emergency disposal task list. A greedy algorithm and DQN with human experience constraints (HEC-DQN) were selected as the benchmark algorithms to conduct a comparative test with HEC-DDQN. Figs.\ref{fig12}(b), (c), and (d) show that the emergency scale was 1. The scheduling method in this paper had the best results according to the three evaluation indicators.

Table \ref{tab-2} summarizes the average values of each index of the three algorithms under different emergency scales, with the best results of each configuration shown in bold.

\begin{table}[htbp]
	\centering
	\caption{Unmanned storage experimental data}
	\resizebox{1.0\linewidth}{!}{
		\begin{tabular}{cp{5.625em}ccc}
			\bottomrule
			\multicolumn{1}{p{5.44em}}{Events' scale} & Algorithm & {$\bar{R}\_{work}$} & {$\bar{rate}_{fail}$} & {$\bar{r}$} \\
			\bottomrule
			\multirow{3}[1]{*}{1} & Greedy & 72.60\% & 53.30\% & {/} \\
			& DQN   & 84.80\% & 24.20\% & -0.563 \\
			& HEC-DDQN & \textbf{88.60\%} & \textbf{11.70\%} & -0.306 \\
			\multirow{3}[0]{*}{2} & Greedy & 60.20\% & 52.00\% & {/} \\
			& DQN   & 78.20\% & 40.80\% & -0.685 \\
			& HEC-DDQN & 85.80\% & 27.30\% & \textbf{-0.217} \\
			\multirow{3}[1]{*}{3} & Greedy & 65.50\% & 56.60\% & {/} \\
			& DQN   & 71.30\% & 38.70\% & -0.586 \\
			& HEC-DDQN & 88.30\% & 23.60\% & -0.31 \\
			\bottomrule
		\end{tabular}%
	}
	\label{tab-2}%
\end{table}%

The table shows the constant 4 emergency events. The emergency disposal effect with the misfire contingency was worse than without, indicating that a fast-developing contingency had a greater impact on the disposal, which is especially noticeable in the greedy algorithm. Moreover, during the experiments, the greedy algorithm took more time to complete, which we believe was caused by recalculating the emergency disposal task list after it was updated. By contrast, the two reinforcement learning algorithms performed better and were more suitable for dealing with such dynamic scheduling problems. In the scale 3 results, the overall emergency effect of the three methods was slightly worse than those from scale 1. We believe the reason for the small gap was the influence of the size of the emergency disposal task list window on the emergency disposal. Taken together, our proposed scheduling method had the best results with the largest mean task completion rate, the smallest mean damage rate of station nodes, and the largest mean value of the reward function.

\subsection{Verification Experiment of Human--Machine Collaborative Emergency Disposal Framework}

\begin{figure}[ht]
	\centering
	\includegraphics[width=8.58cm,height=2.94cm]{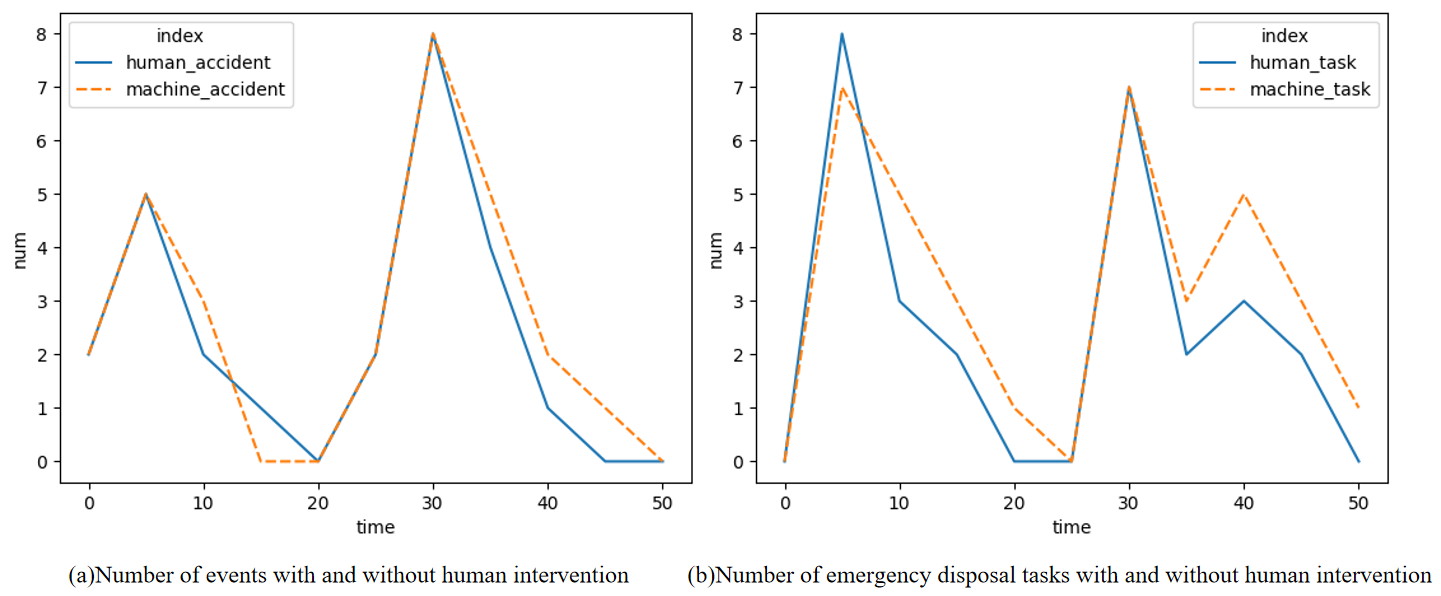}\\
	\caption{Reward functions for different events' scale}
	\label{fig13}
\end{figure}

\begin{figure}[ht]
	\centering
	\includegraphics[width=8.65cm,height=3.00cm]{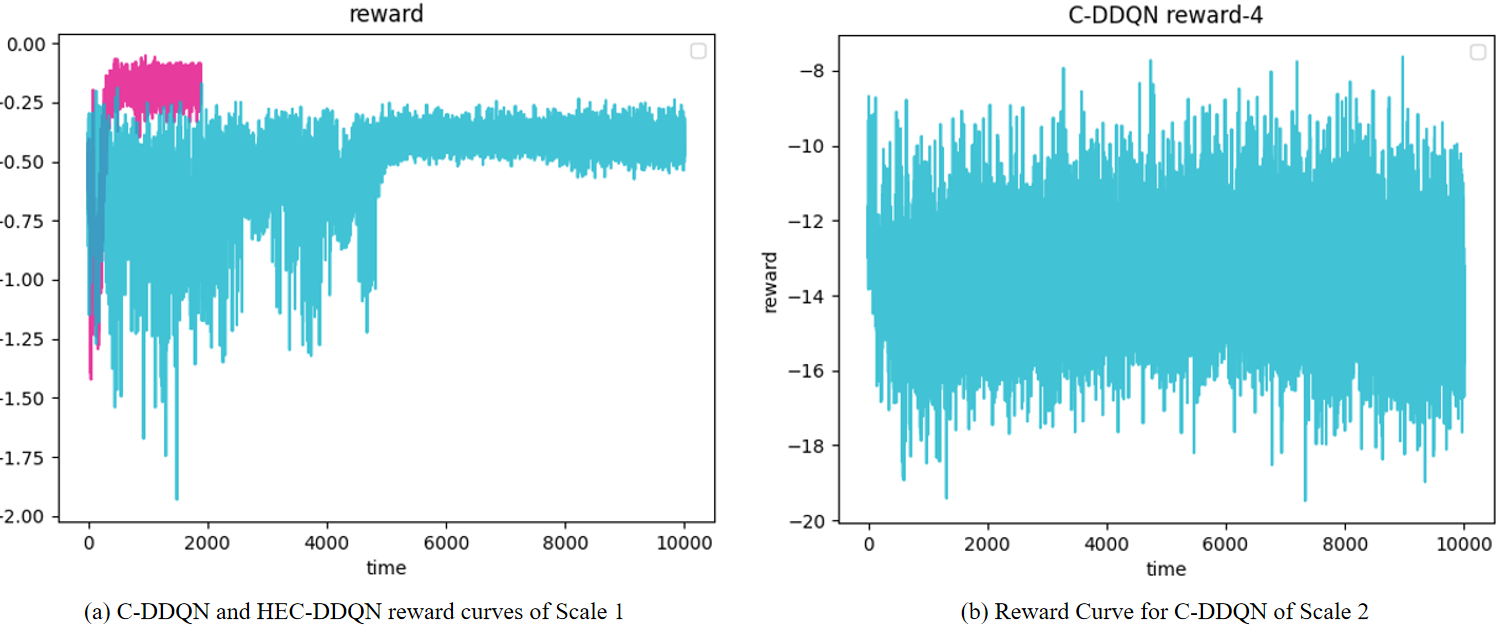}\\
	\caption{Experimental results for the role of human empirical constraints}
	\label{fig14}
\end{figure}

(1)	Analysis of the role of human intervention
Using the two models in the human--machine collaborative emergency disposal method trained in the previous subsections. We used five events appearing at random times and a task window set to 7 as the test scenario. Experiments were conducted under two scenarios of human intervention to modify the emergency disposal task list and unsupervised intervention to use the machine-generated emergency disposal task list directly. Using the number of nodes where emergencies occur, the number of emergency disposal tasks and the emergency response time as indicators, the experimental results are plotted as shown in Fig. \ref{fig14}. Fig. \ref{fig13}(a) shows the situation of the number of events with and without human intervention. The number of emergencies did not differ much, but the machine without human intervention had a somewhat slower decreasing curve of the number of emergencies than the one with human intervention. Fig. \ref{fig13}(b) shows that the number of emergency disposal tasks with human intervention was less than that without human intervention. Therefore, human intervention can reduce the number of emergency disposal tasks with better control of emergencies.

(2)	Analysis of the role of human experience constraints

To test the role of the human experience constraint in the task scheduling stage, we chose to use the classical Double DQN (C-DDQN) without human experience constraints and the HEC-DDQN for emergency disposal task scheduling after obtaining the task list through the task filtering stage in scale 1. The results are shown in Fig. \ref{fig14}. Fig. \ref{fig14}(a) shows the results of the reward functions. HEC-DDQN converged quickly and the reward value was not too low during the trial-and-error process. Fig. \ref{fig15} (b) shows that C-DDQN failed to converge during the testing process in  scale 2. After the comparison experiments, we conclude that DDQN with the human experience constraint we propose better solved the emergency disposal problem in a complex environment.

(3)	Analysis of the effect of human--machine collaborative emergency disposal decision

To verify the effect of the emergency disposal task allocation model as a preliminary task of emergency disposal task scheduling, we selected Double DQN algorithm (C-DDQN) in the phase without Emergency disposal task assignment was selected for comparison with Human--Machine Collaborative Emergency Response Framework (HMC-ERF) proposed here. Similar to the results in the task allocation stage, the emergency disposal task list of the C-DDQN algorithm consisted of emergency nodes and key nodes, with the key nodes obtained by sorting the average value of node centrality commonly used at present. The calculation formula is 
\begin{equation}
	\bar{c}=\frac{1}{c_{b}+c_{o}+c_{i}}.
	\label{eq29}
\end{equation}

\begin{figure}[ht]
	\centering
	\includegraphics[width=7.45cm,height=4.00cm]{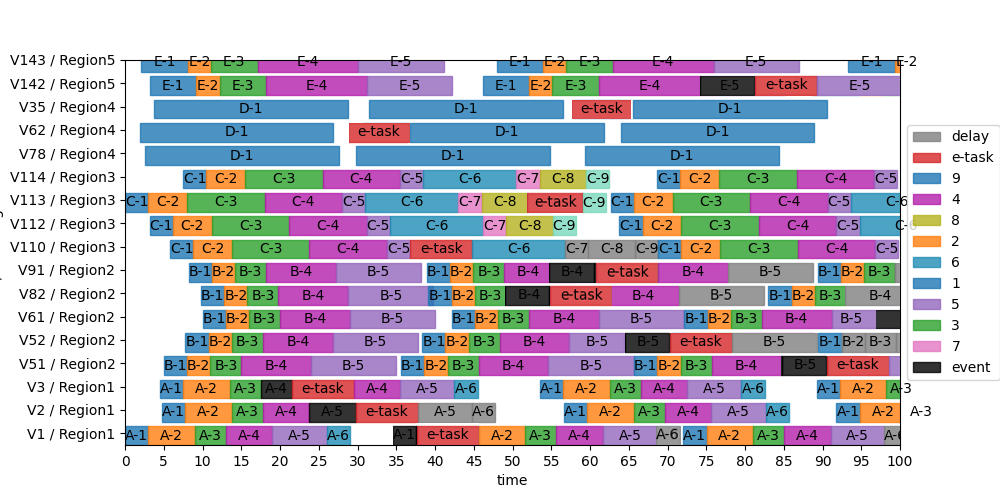}\\
	\caption{Experimental test results of E-DDQN}
	\label{fig16}
\end{figure}

\begin{figure}[ht]
	\centering
	\includegraphics[width=7.45cm,height=4.00cm]{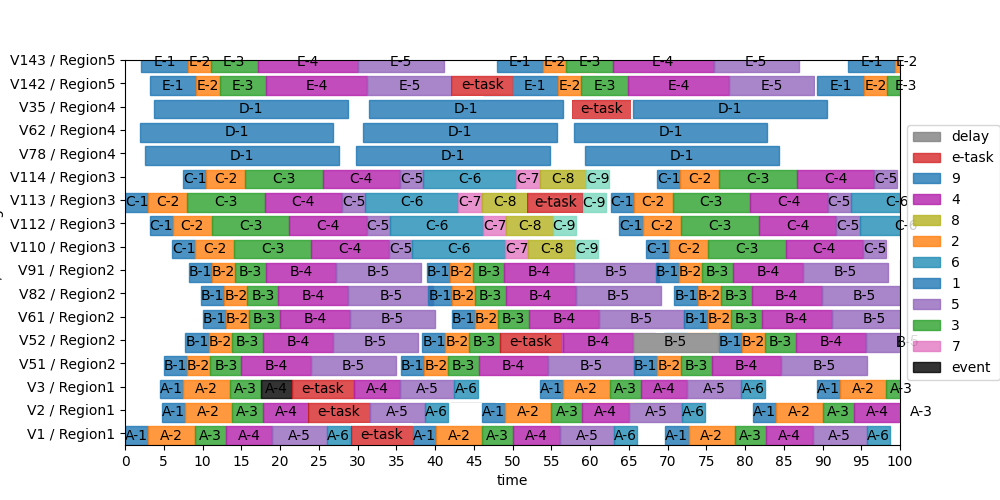}\\
	\caption{Experimental test results of HMC-EDD}
	\label{fig17}
\end{figure}

The two trained models were tested with the emergent event at node V3, and the sequence of operation works and tasks of the relevant nodes is shown in Gantt chart form. Fig. \ref{fig16} shows the operation of E-DDQN and Fig. \ref{fig17} shows the operation of HMC-EDD, where the horizontal axis indicates the time, and the vertical axis indicates the station node number and the region to which it belongs. After the emergent event from node V3, E-DDQN disposed of the emergent node in time to ensure the completion of normal operations, but some of the nodes with potential hazards were not predicted and preventive measures were not taken for them, resulting in more subsequent derivative events. However, it better protected the main traffic nodes and nodes with more operational processes. By contrast, HMC-EDD better controlled the spread of emergency events and delayed fewer normal work steps.
We conclude that the man--machine collaborative emergency disposal method we have proposed better predicted the development of emergencies, effectively controlled the spread of emergency events, and improved the efficiency of emergency disposal.

\section{Conclusion}

Emergency events occur frequently in modern operating environments, including those with a complexity that makes them suitable for unmanned systems. However, human trust in unmanned systems is not high at present. To solve this contradiction, we propose a man--machine collaborative emergency disposal method. It is divided into task filtering and task scheduling stages and adopts different human--computer interaction methods to solve the problem of emergency disposal decision-making with as little disruption of normal operations as possible. Complex operational situations are simplified by graph calculation, and the complete process of emergency disposal is abstracted into filtering key nodes in the graph with the task nodes for emergency disposal operations inserted into the procedure subgraph.

\end{document}